\documentclass[lettersize,journal]{IEEEtran}
\usepackage{amsmath,amsfonts}
\usepackage{algorithmic}
\usepackage{algorithm}
\usepackage{array}
\usepackage[caption=false,font=normalsize,labelfont=sf,textfont=sf]{subfig}
\usepackage{textcomp}
\usepackage{stfloats}
\usepackage{url}
\usepackage{verbatim}
\usepackage{graphicx}
\usepackage{cite}

\usepackage[dvipsnames]{xcolor}
\definecolor{aliceblue}{rgb}{0.94, 0.97, 1.0}
\usepackage{circledsteps}
\usepackage{graphicx}
\usepackage{float}
\usepackage{stfloats}
\usepackage{multirow}
\usepackage{booktabs}
\usepackage{amssymb}
\usepackage{bbding}
\usepackage{pifont}
\usepackage{wasysym}
\usepackage{utfsym}
\usepackage{fontawesome}
\usepackage{soul} 
\usepackage{color, xcolor} 
\usepackage{colortbl} 
\usepackage{xcolor}
\usepackage{array}  
\definecolor{Color-High-Level}{HTML}{BEDAE8}
\definecolor{Color-Low-Level}{HTML}{CCE0CA}

\begin{document}

\title{ReaMOT: A Benchmark and Framework for Reasoning-based Multi-Object Tracking}

\author{Sijia Chen$^{\ast}$, Yanqiu Yu$^{\ast}$, En Yu$^{\ast}$, and Wenbing Tao,~\IEEEmembership{Member,~IEEE}
\thanks{$^{\ast}$ Equal contribution.}
\thanks{This work was supported by the National Natural
 	Science Foundation of China under Grant 62576144.}
\thanks{\IEEEcompsocthanksitem Sijia Chen, Yanqiu Yu, En Yu, and Wenbing Tao are with the National Key Laboratory of Science and Technology on Multi-spectral Information Processing, School of Artificial Intelligence and Automation, Huazhong University of Science and Technology, Wuhan 430074, China. E-mail: \{sijiachen, yanqiuyu6, yuen, wenbingtao\}@hust.edu.cn. (Corresponding author: Wenbing Tao)}
}

%
\markboth{XXX,~Vol.~XX, No.~XX, MAY~2026}%
{Shell \MakeLowercase{\textit{et al.}}: A Sample Article Using IEEEtran.cls for IEEE Journals}


\maketitle

\begin{abstract}
Referring Multi-Object Tracking (RMOT) aims to track targets specified by language instructions. However, existing RMOT paradigms heavily rely on explicit visual-textual matching and consequently fail to generalize to complex instructions that require logical reasoning. To overcome this, we propose \textbf{Rea}soning-based \textbf{M}ulti-\textbf{O}bject \textbf{T}racking (\textbf{ReaMOT}), a novel task that elevates tracking to a cognitive level, requiring models to infer and track specific targets satisfying implicit constraints via logical reasoning. To advance this field, we construct the \textbf{ReaMOT Challenge}, a comprehensive benchmark featuring a tailored metric suite and a large scale dataset. This dataset comprises 1,156 language instructions, 423,359 image language pairs, and 869 distinct video sequences systematically categorized into six distinct evaluation scenarios, with over 75\% of the instructions dedicated to High Level Reasoning. Furthermore, recognizing that traditional trackers lack cognitive capacity while direct application of Large Vision-Language Model (LVLM) yields severe temporal inconsistencies, we propose \textbf{ReaTrack}. Driven by the insight to decouple high-level cognitive localization from low-level physical motion continuity, this training-free framework dynamically aligns the semantic detections of a Thinking-variant LVLM with the robust motion priors of SAM2. Extensive experiments on the ReaMOT Challenge benchmark demonstrate that ReaTrack establishes a new leading performance standard. Notably, it achieves a more than threefold improvement in RHOTA on the High Level Reasoning subset. Our dataset and code will be available at \textbf{https://github.com/chen-si-jia/ReaMOT}.

\end{abstract}



\begin{figure}[t]
\centering
    \includegraphics[width=1.0\linewidth]{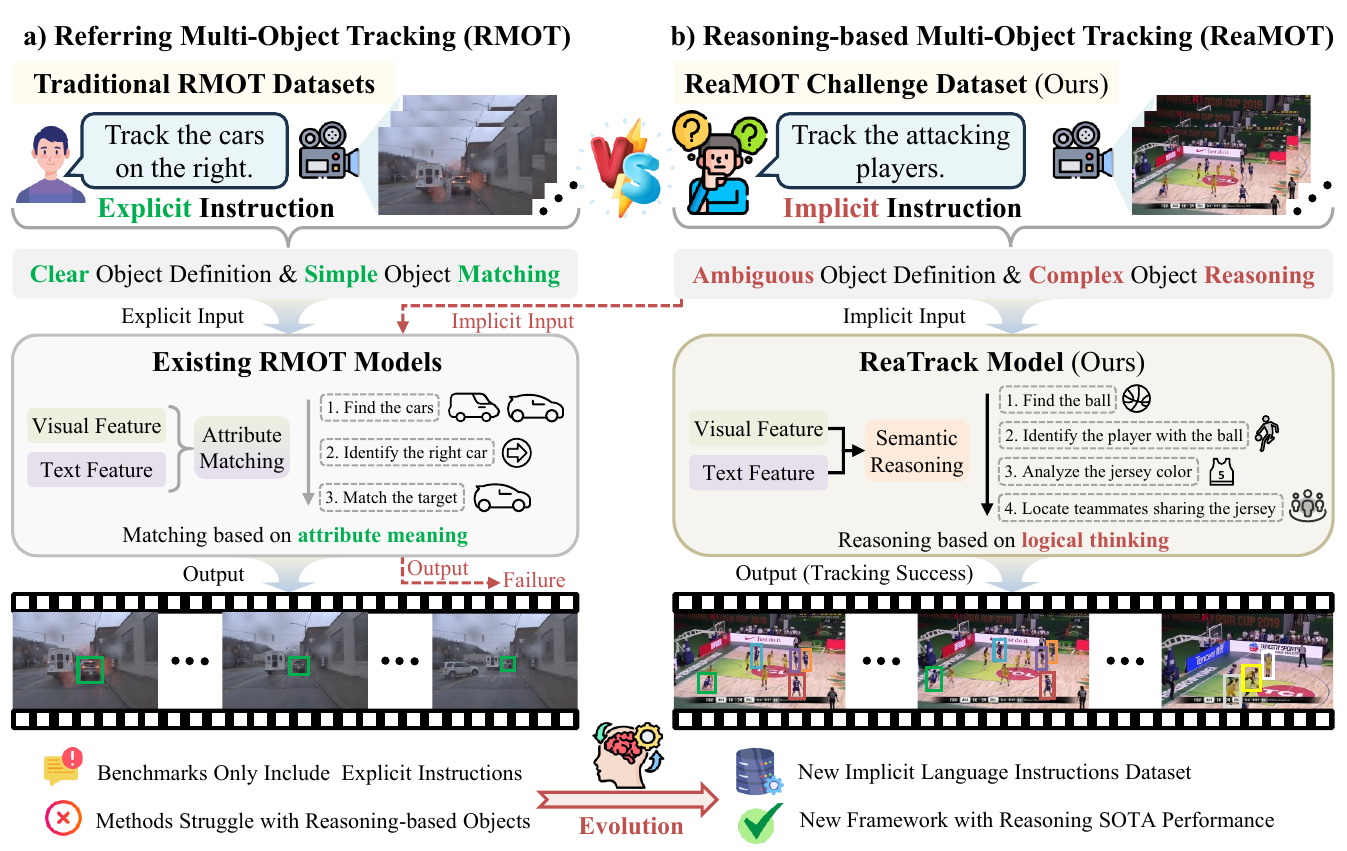}
    \caption{\textbf{Comparison between the traditional RMOT and the proposed ReaMOT tasks.} \textbf{(a) The traditional RMOT task} relies on explicit visual attribute matching (e.g., “car”, “right”). Consequently, existing models designed for this standard paradigm fail when confronted with implicit language instructions. \textbf{(b) In contrast, the ReaMOT task} elevates the tracking process to a cognitive level. It requires models to infer and track the specific targets that satisfy implicit semantic constraints (e.g., “attacking players”) through multiple steps of logical reasoning.}
    \label{fig:RMOT&ReaMOT.}
\end{figure}

\section{Introduction}

Multi-Object Tracking (MOT) stands as a fundamental task in computer vision, serving critical applications such as video surveillance \cite{yi2024ucmctrack}, autonomous driving \cite{zhang20233d}, and intelligent transportation \cite{bashar2022multiple}. While generic MOT methods have achieved remarkable success in purely visual scenarios, integrating linguistic cues remains a significant challenge. To address this, the task of Referring Multi-Object Tracking (RMOT) \cite{wu2023referring} was recently introduced, requiring models to track targets specified by language instructions.

However, existing RMOT paradigms \cite{wu2023referring, zhang2024bootstrapping, du2024ikun, chen2025cross, li2025lamot, li2025language, liang2025cognitive} are largely designed for explicit instructions (e.g., “Track the cars on the right.”) and consequently fail to generalize to complex instructions that require logical reasoning. In real-world scenarios, users rarely rely on simple adjectives alone. Instead, they refer to targets through complex semantic cues involving relationships, behaviors, and causal intentions. For example, in crowded scenes with dense objects, single attributes are often insufficient to identify the intended targets. In contrast, humans can naturally resolve such ambiguities by reasoning over contextual information and inferring implicit intentions from prior knowledge.

Motivated by these observations, we propose a novel task, named \textbf{Rea}soning-based \textbf{M}ulti-\textbf{O}bject \textbf{T}racking (\textbf{ReaMOT}). Unlike standard RMOT, ReaMOT elevates the tracking process to a cognitive level. As illustrated in Fig.~\ref{fig:RMOT&ReaMOT.} (a), the traditional RMOT relies strictly on explicit visual attribute matching (e.g., “car”, “right”). Consequently, existing models designed for this standard paradigm intrinsically fail when confronted with implicit language instructions. In contrast, as shown in Fig.~\ref{fig:RMOT&ReaMOT.} (b), the ReaMOT task requires models to infer and track the specific targets that satisfy implicit semantic constraints. For instance, identifying “attacking players” requires the model to execute a sequential reasoning process that involves locating the ball, determining active possession, extracting the jersey color of that specific player, and finally generalizing this visual attribute to successfully locate all corresponding teammates.

To advance research in this direction, we construct the \textbf{ReaMOT Challenge}, a comprehensive benchmark serving as a rigorous evaluation platform. The benchmark features a large-scale dataset comprising 1,156 language instructions with reasoning characteristics, 423,359 image language pairs, and 869 distinct video sequences spanning 21 target categories. To ensure high data quality, we adopt a hybrid annotation pipeline that combines GPT-assisted feature analysis with rigorous manual verification. Distinct from conventional datasets, our language annotations are heavily characterized by a reasoning-driven vocabulary that captures complex social relationships, underlying intents, and probabilistic uncertainties. Furthermore, we establish a set of reasoning level classification criteria to enable fine grained evaluation. Based on these criteria, the language instructions in the dataset are categorized into two subsets. The \textit{High Level Reasoning} subset contains 868 instructions, and the \textit{Low Level Perception} subset contains 288 instructions, ensuring thorough coverage of varying cognitive depths. To comprehensively evaluate model generalization, these instructions are categorized into six distinct evaluation scenarios, including Open World and Autonomous Driving, providing highly demanding visual foundations. Finally, we propose a tailored metric suite designed to quantify both reasoning accuracy and tracking robustness, facilitating a holistic evaluation of model capabilities.

To tackle the ReaMOT task, current RMOT paradigms face severe bottlenecks. Traditional tracking architectures lack the cognitive capacity to decode implicit logical deductions. Simultaneously, while recent Large Vision-Language Models (LVLMs) \cite{liu2024visual, chen2024internvl, bai2025qwen2, bai2025qwen3} exhibit advanced reasoning abilities, directly applying them for tracking yields severe temporal inconsistencies and trajectory fragmentation because they inherently lack video level motion priors.  Driven by this insight, we argue that an effective reasoning-based tracker must decouple high-level cognitive localization from low-level physical motion continuity. To this end, we propose \textbf{ReaTrack}, a strong training-free baseline framework. It comprises three synergistic modules: (1) \textit{Reasoning-Aware Detection (RAD)}, which leverages a Thinking-variant LVLM to resolve semantic ambiguities and provide highly reliable semantic detections; (2) \textit{Mask-Based Temporal Propagation (MBTP)}, which employs SAM2 \cite{ravi2024sam} to overcome the temporal inconsistency of LVLMs by generating robust physical motion priors; and (3) \textit{Reasoning-Motion Association (RMA)}, a core decision-making unit that dynamically aligns cognitive detections with temporal predictions to robustly manage trajectory lifecycles.

Finally, we evaluate the ReaTrack framework on the ReaMOT Challenge benchmark under zero-shot settings. Extensive experiments demonstrate that ReaTrack establishes a new leading performance standard by consistently outperforming existing methods. Notably, in the High Level Reasoning subset, ReaTrack achieves an RHOTA of 42.28\%, delivering a more than threefold improvement over the best competing method. Furthermore, it achieves top scores across all metrics, fully validating its superior capability in generalizing from basic visual recognition to decoding implicit instructions within complex reasoning scenarios.

In summary, our main contributions are as follows:
\begin{itemize}
    \item We propose \textbf{Rea}soning-based \textbf{M}ulti-\textbf{O}bject \textbf{T}racking (\textbf{ReaMOT}), a novel and challenging task that elevates tracking to a cognitive level, requiring models to infer and track specific targets satisfying implicit constraints via logical reasoning.
    \item We construct the \textbf{ReaMOT Challenge}, a comprehensive benchmark comprising a tailored metric suite and a large scale dataset. It features 1,156 language instructions and 869 distinct video sequences systematically categorized into \textbf{six distinct evaluation scenarios}, with over \textbf{75\%} of the instructions dedicated to \textit{\textbf{High Level Reasoning}}.
    \item We propose \textbf{ReaTrack}, a training-free framework that synergizes the high-level cognitive reasoning capabilities of Thinking-LVLM with the low-level physical temporal propagation of SAM2.
    \item Extensive experiments demonstrate that the ReaTrack framework achieves \textbf{state-of-the-art} performance. Remarkably, it delivers over a \textbf{threefold improvement} in RHOTA on the challenging High-Level Reasoning subset, thereby establishing a strong baseline for future research.
\end{itemize}

\section{Related Work}
\label{sec:Related_Work}

\subsection{Referring Multi-Object Tracking}
Referring Multi-Object Tracking (RMOT) extends traditional Multi-Object Tracking (MOT) \cite{zhang2021fairmot, zhang2022bytetrack, zeng2022motr, chen2024delving, gao2025multiple, ma2025multi, huang2025dftrack, xue2025usvtrack, guo2026occlusion} by incorporating natural language to selectively track specific objects. Unlike generic MOT, RMOT demands robust joint vision-language understanding to localize targets while maintaining consistent identities over time. Following the pioneering end-to-end framework, TransRMOT \cite{wu2023referring}, subsequent research has progressively refined this paradigm. To enhance trajectory continuity and computational efficiency, TempRMOT \cite{zhang2024bootstrapping} introduces temporal context modeling, while iKUN \cite{du2024ikun} streamlines the architecture for efficient deployment. Expanding upon this temporal foundation, TenRMOT \cite{xiao2025temporal} integrates rich historical cues to unify both robust tracking and fine grained pixel level segmentation. Furthermore, recent advancements have expanded RMOT to tackle more complex real world constraints. For instance, CRTracker \cite{chen2025cross} leverages multi camera observations to mitigate severe occlusions. Concurrently, to achieve deeper cross modal alignment, models such as DKGTrack \cite{li2025language} and CDRMT \cite{liang2025cognitive} focus on semantic guidance and the disentanglement of appearance, motion, and language features. Furthering this direction, CGATracker \cite{zhuang2025cgatracker} introduces correlation aware graph alignment to establish reliable multimodal associations, bolstering overall tracking robustness.

To support these methods, several datasets provide dedicated benchmarks. Refer-KITTI \cite{wu2023referring} and its extension Refer-KITTI-V2 \cite{zhang2024bootstrapping} establish benchmarks for autonomous driving scenarios. Subsequent datasets expand interaction diversity and viewing domains, including Refer-Dance \cite{du2024ikun} for highly dynamic human movements, LaMOT \cite{li2025lamot} for long-term trajectory associations, and AerialMind \cite{chen2026aerialmind} for Unmanned Aerial Vehicle (UAV) scenarios. However, existing datasets and methods primarily focus on explicit visual perception. They heavily rely on direct language descriptions of physical attributes (e.g., color, location) and struggle to process implicit instructions demanding logical deductions, social common sense, or intention prediction. Our work addresses this gap, upgrading RMOT from basic perception to advanced reasoning.

\subsection{Large Vision-Language Models}
Large Vision-Language Models (LVLMs) build unified multimodal representations by jointly training on image and text data. Driven by Large Language Models (LLMs), recent architectures integrate powerful visual encoders with LLMs via cross-modal projectors, shifting away from early contrastive learning paradigms. Pioneering models like LLaVA \cite{liu2024visual} demonstrate the potential of visual instruction tuning. Subsequently, InternVL \cite{chen2024internvl} captures fine-grained visual semantics, and the Qwen-VL series \cite{bai2025qwen2, bai2025qwen3} significantly advances spatial-semantic grounding and temporal understanding. Through these advancements, modern LVLMs demonstrate remarkable emergent capabilities, including zero-shot generalization, rich world knowledge, and complex logical deduction, which serve as the critical foundation for reasoning-based perception tasks.

Empowered by these capabilities, LVLMs expand from explicit visual recognition into reasoning-driven perception. For instance, DetGPT \cite{pi2023detgpt} locates targets based on user intentions, TrackGPT \cite{zhu2023tracking} comprehends intentions for single-object tracking, and VRS-HQ \cite{gong2025devil} enhances temporal reasoning for video segmentation. Despite these strides, utilizing complex language reasoning for tracking multiple dynamic targets remains largely unexplored, as existing methods focus on single targets and explicit cues. To fill this gap, we construct the ReaMOT Challenge Benchmark, providing a comprehensive platform to evaluate multimodal models in complex reasoning-based multi-object tracking scenarios.

\section{Benchmark}
\label{sec:Benchmark}

To advance the research on the reasoning-based multi-object tracking (ReaMOT) task and evaluate the models' reasoning and tracking abilities, we construct a reasoning-based multi-object tracking benchmark, named ReaMOT Challenge. Below, we provide details about this benchmark.

\subsection{Dataset Construction}

\subsubsection{Dataset Collection}
\label{subsubsec:Dataset Collection}
Our ReaMOT task requires language instructions with inherent reasoning properties, where target identification relies on a series of semantic and logical inferences. In addition, the task demands data presented in video format and involving multiple targets. Based on these requirements, the source datasets used to construct the ReaMOT Challenge must simultaneously satisfy three criteria: target attributes that support deep reasoning, multi-object scenes, and video sequence. Accordingly, we select and integrate 12 publicly available datasets that meet these criteria, including Argoverse-HD \cite{li2020towards}, DanceTrack \cite{sun2022dancetrack}, GMOT-40 \cite{bai2021gmot}, KITTI \cite{geiger2012we}, MOT17 \cite{milan2016mot16}, MOT20 \cite{dendorfer2020mot20}, MPHOI \cite{qiao2022geometric}, PathTrack \cite{manen2017pathtrack}, PoseTrack \cite{andriluka2018posetrack}, SportsMOT \cite{cui2023sportsmot}, UA-DETRAC \cite{wen2020ua}, and UAVDT \cite{du2018unmanned}. These datasets consist of video sequences containing at least two targets, featuring complex scenes and rich target characteristics, and are therefore well suited for reasoning-based multi-object tracking task. Overall, these source datasets cover 21 target categories, such as people, vehicles, birds, and balls. Detailed statistics and characteristics of these source datasets are summarized in Table~\ref{tab:Source datasets of the ReaMOT Challenge benchmark.}.

\begin{table}[t]
\setlength{\abovecaptionskip}{0.6mm}
\caption{\textbf{Source datasets of the ReaMOT Challenge benchmark.}}
\label{tab:Source datasets of the ReaMOT Challenge benchmark.}
\resizebox{1.0 \columnwidth}{!}{
    \centering
    \begin{tabular}{l|l|l}
        \toprule
        \rowcolor[HTML]{BFDAE9}
        \textbf{Dataset} & \textbf{Object Category} & \textbf{Characteristics that make sense for the ReaMOT Challenge benchmark} \\
        \midrule
        \multirow{3}*{Argoverse-HD \cite{li2020towards}} & bicyle, bus, motorcycle, & \multirow{3}*{complex real-world scenes, multi-category transportation item annotation} \\
         & car, person, stop sign, &  \\
         & traffic light, truck &  \\
        \midrule
        DanceTrack \cite{sun2022dancetrack} & dancer & similar appearance, complex movement \\
        \midrule
        \multirow{3}*{GMOT-40 \cite{bai2021gmot}} & airplane, ball, balloon, & \multirow{3}*{diversity of target categories, real-world challenges} \\
         & bird, boat, car, fish, &  \\
         & insect, person, stock &  \\
        \midrule
        KITTI \cite{geiger2012we} & car, person & multiple scenes, multiple categories \\
        \midrule
        MOT17 \cite{milan2016mot16} & person & diverse scenes \\
        \midrule
        MOT20 \cite{dendorfer2020mot20} & person & high pedestrian density \\
        \midrule
        MPHOI \cite{qiao2022geometric} & person & multiplayer interactive activities \\
        \midrule
        PathTrack \cite{manen2017pathtrack} & pedestrian & different complex application scenarios (weather, time, location, etc.) \\
        \midrule
        PoseTrack \cite{andriluka2018posetrack} & person & a variety of daily activities and sports \\
        \midrule
        \multirow{2}*{SportsMOT \cite{cui2023sportsmot}} & \multirow{2}*{athlete} & diverse sports scenes, with fast and variable speed movements \\
         & & and similar but distinguishable looking goals \\
        \midrule
        UA-DETRAC \cite{wen2020ua} & car & real-world traffic scenarios (weather, lighting conditions) \\
        \midrule
        UAVDT \cite{du2018unmanned} & car & real-world traffic scenarios (weather, lighting conditions) \\
        \bottomrule
    \end{tabular}
}
\end{table}

\begin{table}[t]
\setlength{\abovecaptionskip}{0.6mm}
\caption{\textbf{Attribute criteria for the ReaMOT Challenge benchmark.}}
\label{tab:Appendix Attribute criteria for the ReaMOT Challenge benchmark.}
\resizebox{1.0\linewidth}{!}{
    \centering
    \begin{tabular}{l|l|l}
    \toprule
    \rowcolor[HTML]{BFDAE9}
    \textbf{Attribute} & \textbf{Detailed Attribute} & \textbf{Example} \\
    \midrule
    \multirow{3}{*}{\textbf{Spatial Position}} 
    & orientation & right; at the front. \\
    
    & long time position change & from right to left. \\
    
    & relative position & closest to the curb; on the opposite side of the river \\
    \midrule
    \multirow{6}{*}{\textbf{Movement}} 
    & concrete movement & crouching; sitting down. \\
    
    & generalized behavior & difficult move; violent act. \\
    
    & movement tendency & driving farther and farther. \\
    
    & social behavior & ribbon-cutting event. \\
    
    & multi-person associated action & walking arm in arm. \\
    
    & action modifier & quick; agile; intently. \\
    \midrule
    \multirow{3}{*}{\textbf{Costume}} 
    & color & green; brightly colored. \\
    
    & style & overalls; wearing cool clothes. \\
    
    & modifier & with the same texture as zebras; dressed up like a butterfly. \\
    \midrule
    \multirow{8}{*}{\textbf{Human Attribute}} 
    & gender & girl; boy. \\
    
    & age & older; young. \\
    
    & manner & smiling; happy; showing disgust. \\
    
    & appearance & tall; pale skin; light hair. \\
    
    & figure & slim; well-shaped; short man. \\
    
    & personality & loves baseball; queenly demeanor; confident and eager to perform. \\
    
    & mental activity & focused on tennis; desire to explore; striving to win. \\
    
    & mood & happy; smiling with sadness; controlling emotions. \\
    \midrule
    \multirow{3}{*}{\textbf{Object Attribute}} 
    & color & red; exceptional color. \\
    
    & size & large; small. \\
    
    & usage & used mainly for transporting goods. \\
    \midrule
    \multirow{2}{*}{\textbf{Specific Noun}} 
    & direct description & gymnasts; athletes; leaders. \\
    
    & figurative description & background panel. \\
    \midrule
    \multirow{3}{*}{\textbf{Auxiliary Modifier}} 
    & adjective & passionate; eye-catching. \\
    
    & adverb & very loudly; conveniently. \\
    
    & time determiner & when it is dark; at the start of the game. \\
    \midrule
    \multirow{4}{*}{\textbf{Others}} 
    & interrelation & family members; smiled at each other, sharing a mutual understanding. \\
    
    & aim & to transport goods; waiting for passengers; maintaining composure. \\
    
    & common sense interpretation & public transport vehicles; environmentally friendly transportation tools \\
    
    & event trend words & more likely to play football; disrupt the smooth flow of following vehicles. \\
    \bottomrule
    \end{tabular}
}
\vspace{-1pt}
\end{table}

\subsubsection{Attribute Criteria}
\label{subsubsec:Attribute Criteria}
To ensure the comprehensiveness and standardization of language attributes in the dataset, the annotators summarize a highly representative attribute criteria based on extensive research in human language literature. Considering the infinite nature of human language, the aim is not to exhaust all words, but rather to establish an attribute space that covers the core semantic dimensions. This criteria systematically organizes linguistic dimensions into eight main categories, including spatial position, movement, costume, human attribute, object attribute, specific noun, auxiliary modifier, and others, thereby covering the core attributes of language instructions. The primary motivation for developing this criteria is to provide explicit and rich attribute references for GPT-based feature generation during dataset construction. By predefining these dimensions, the model is guided to produce more targeted content and to avoid overly random generation that may result in limited or unbalanced descriptive coverage. Through this approach, the generated features are able to capture not only observable physical appearances and behaviors, but also higher-level intentions and social actions. In addition, we ensure mutual exclusivity among the major categories while allowing flexible expressions within each attribute, thereby accommodating the diversity of real-world language. The attribute criteria are presented in Table~\ref{tab:Appendix Attribute criteria for the ReaMOT Challenge benchmark.}.

\subsubsection{Data Annotation}
To construct high-quality language instructions with reasoning characteristics, we organize a three-group annotation team and evenly divide all videos among them for independent annotation. After completing the initial annotations, the annotators cross-check each other’s work to ensure accuracy and consistency. All annotation tasks strictly follow the procedure outlined below. First, each annotator thoroughly reviews the video sequences and preliminarily selects several target objects for annotation. Next, key frames containing the selected targets are manually extracted, and, guided by the attribute criteria, GPT-4o \cite{hurst2024gpt} is employed to generate descriptive features of the specified targets. Finally, the annotators verify the accuracy and validity of the GPT-4o outputs and integrate them into complete language instructions. The annotation process is shown in Fig.~\ref{fig:Dataset Annotation Process.}. The specific annotation process is as follows:

\begin{figure*}[t]
\setlength{\abovecaptionskip}{1.2mm}
\centering
    \includegraphics[width=1.0\linewidth]{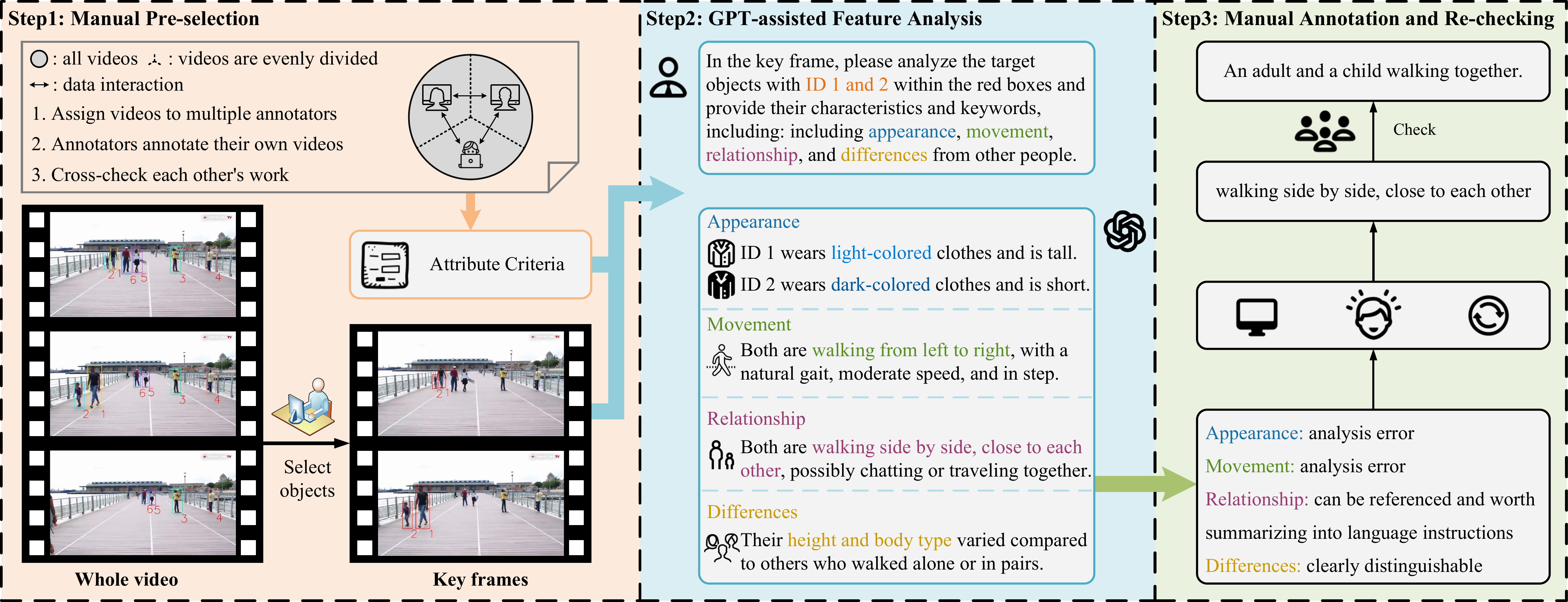}
    \caption{\textbf{Overview of the annotation pipeline.} The process comprises three stages to yield high-quality, reasoning-based instructions: (1) \textbf{Manual Pre-selection}: Annotators review video sequences to select key frames and target objects that share common features yet exhibit distinctive attributes, guided by our \textit{Attribute Criteria}; (2) \textbf{GPT-assisted Feature Analysis}: Key frames overlaid with bounding boxes (bbox) and identity (ID) numbers are fed into GPT. Guided by structured prompts, the model generates comprehensive descriptive features covering multi-dimensional attributes (e.g., appearance, motion, relationships, and intention); (3) \textbf{Manual Annotation and Re-checking:} Annotators verify the GPT outputs, filter out inaccurate analysis, and synthesize the validated characteristics into final language instruction with reasoning characteristics.}
    \label{fig:Dataset Annotation Process.}
    \vspace{-5.8pt}
\end{figure*}

\textbf{Step 1: Manual Pre-selection.} First, we visualize the annotations from the source dataset on the images. Then, we manually review the entire video. Finally, we select key frames that contain multiple objects suitable for language instructions with reasoning characteristics. Our core selection criterion dictates that target objects must share foundational characteristics to create visual ambiguity, yet exhibit reasoning-dependent attributes (e.g., tactical roles, social interactions, and behavioral intents). These features provide a foundation for annotating language instructions with reasoning characteristics.

\textbf{Step 2: GPT-assisted Feature Analysis.} 
To improve efficiency while ensuring accuracy, we employ GPT-4o to perform preliminary filtering and content analysis of the images. Specifically, we first select key frames annotated with bounding boxes (bbox) and identity (ID) numbers as input, which are then provided to GPT-4o. Based on this input, and guided by attribute criteria (introduced in Section~\ref{subsubsec:Attribute Criteria}) and specific prompts, GPT-4o generates descriptive features of the target objects within the key frames.

\textbf{Step 3: Manual Annotation and Re-checking.} The characteristics provided by GPT-4o \cite{hurst2024gpt} are summarized and generalized manually. After manual review, the final language instructions with reasoning characteristics are obtained. These instructions can cover simple features such as the targets' appearance or motion characteristics, as well as more complex attributes like human cognitive activities or vehicle movement trends. It is important to note that these instructions require deep reasoning to analyze and determine which targets they describe. For example, the instruction “The people in the scene are close, possibly family members.” requires reasoning to determine which individuals in the video are close to each other, and may have a familial relationship.

Following the above procedure, we construct the ReaMOT Challenge dataset. A critical feature of the dataset is that target annotations are spatiotemporally dynamic. Rather than fixing identities based solely on initial appearance, the ground-truth bounding boxes strictly follow the evolving semantic state defined by the language instructions. This dynamic protocol can be illustrated in complex scenarios. For instance, given the prompt “track the attacking players.” in a basketball sequence, annotations exclusively correspond to the team currently holding ball possession. When a turnover or defensive rebound occurs, the semantic state shifts immediately. As a result, trajectories of the previous offensive players are terminated, and new attacking players are dynamically initialized. In this way, the annotated targets remain strictly and continuously aligned with the language instructions throughout the video.

\begin{table*}[t]
\setlength{\abovecaptionskip}{0.6mm}
    \centering
    \caption{\textbf{Core attribute domains at different reasoning levels.} Representative attribute domains, descriptions, and examples for High-Level Reasoning and Low-Level Perception.}
    \label{tab:Core Attribute Domains}
    \resizebox{1.0 \textwidth}{!}{
        \begin{tabular}{l|l|l|l}
        \toprule
        \rowcolor[HTML]{BFDAE9}
        \textbf{Reasoning Level} & \textbf{Attribute Domain} & \textbf{Description} & \textbf{Examples} \\
        \midrule
        \multirow{5}{*}{High-Level Reasoning} 
        & Functional Space 
        & Social rules or functional definitions associated with physical areas. 
        & intersection, bicycle lane, no-parking zone, office area. \\
        
        & Social Label 
        & Social identity, occupation, or specific role of an individual. 
        & police officer, athlete, doctor, lead dancer, suspect. \\

        & Psychological Intention 
        & Inference of internal emotions, motivations, or immediate intentions. 
        & nervous, prepared, angry, attempting to escape, hostile. \\

        & Causal Logic 
        & Reasoning about underlying causes, purposes, or causal chains. 
        & due to, in order to, possibly because, resulting in. \\

        & Social Attributes 
        & Social conventions, cultural values, or rule boundaries.
        & expensive, professional, illegal, foul, yielding. \\
        \midrule
        \multirow{5}{*}{Low-Level Perception} 
        & Basic Posture 
        & Instinctive biological actions observable through direct perception. 
        & standing, squatting, walking, running, raising hands, rotating. \\

        & Physical Attributes 
        & Primitive material properties of objects or humans. 
        & color, size, shape, brightness contrast, texture, material. \\

        & Spatial Coordinates 
        & Absolute and relative physical positions in space. 
        & left, center, right, foreground, background, distance, depth. \\

        & Motion Trajectory 
        & Physical movement paths of dynamic objects. 
        & turning, stopping, linear movement, acceleration. \\

        & Biological Traits 
        & Basic human physiological identifiers. 
        & gender, approximate age. \\
        \bottomrule
        \end{tabular}
    }
    \vspace{-3pt}
\end{table*}

\begin{figure*}[t]
\setlength{\abovecaptionskip}{1.2mm}
\centering
\subfloat[\label{fig:a}]{
    \includegraphics[width=0.30\linewidth]{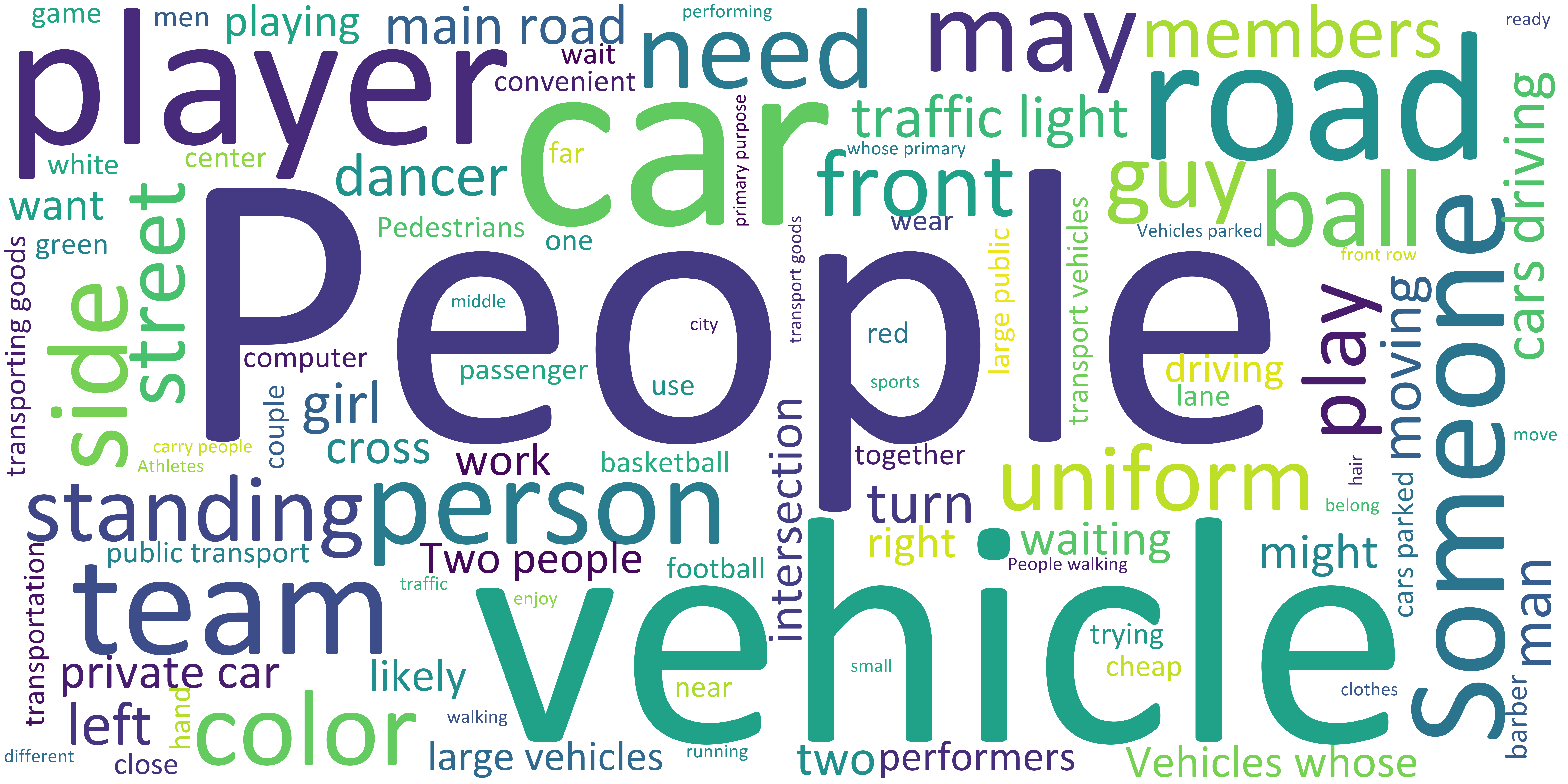}}
\hfil
\subfloat[\label{fig:b}]{
    \includegraphics[width=0.30\linewidth]{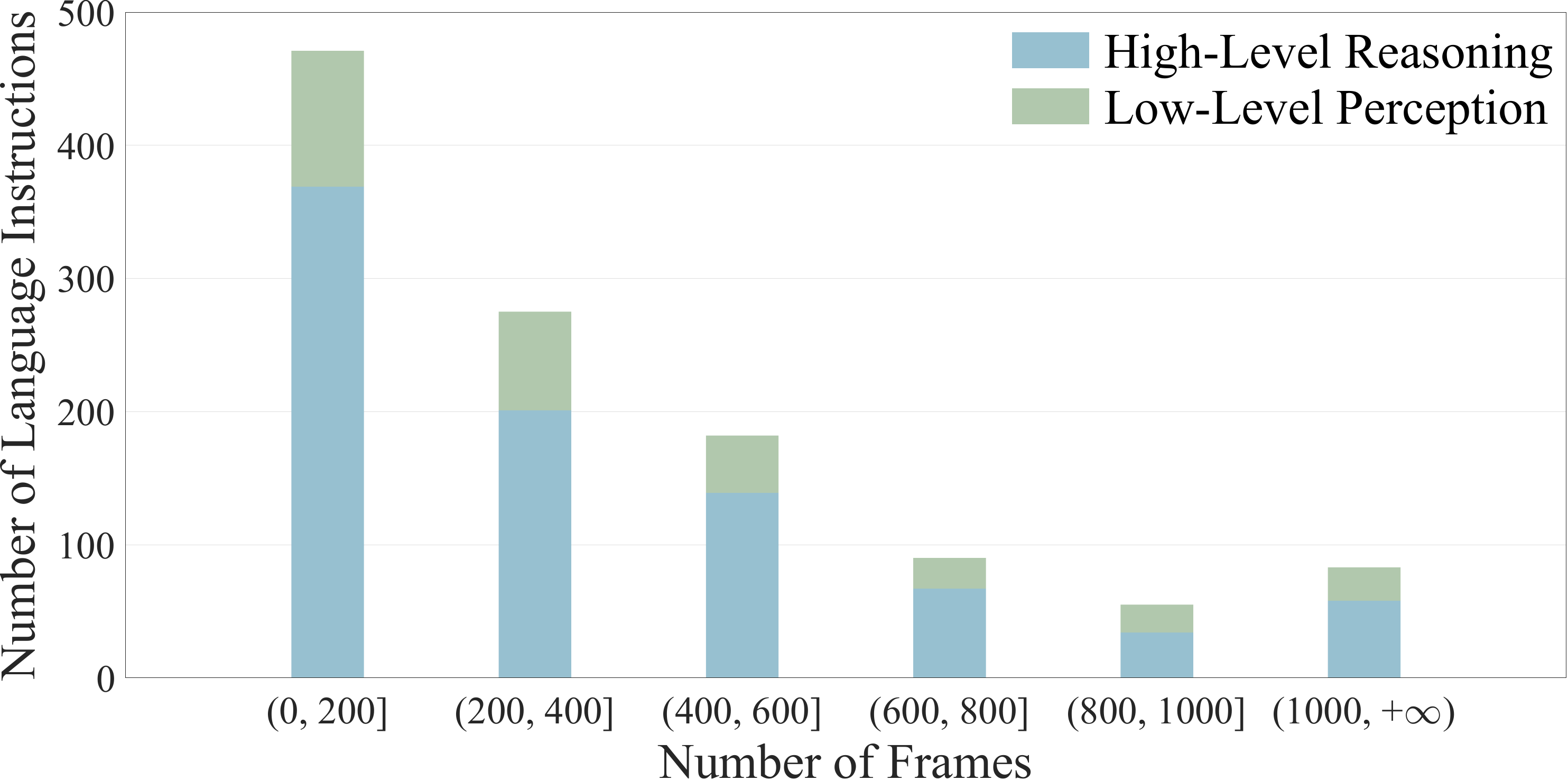}}
\hfil
\subfloat[\label{fig:c}]{
    \includegraphics[width=0.183\linewidth]{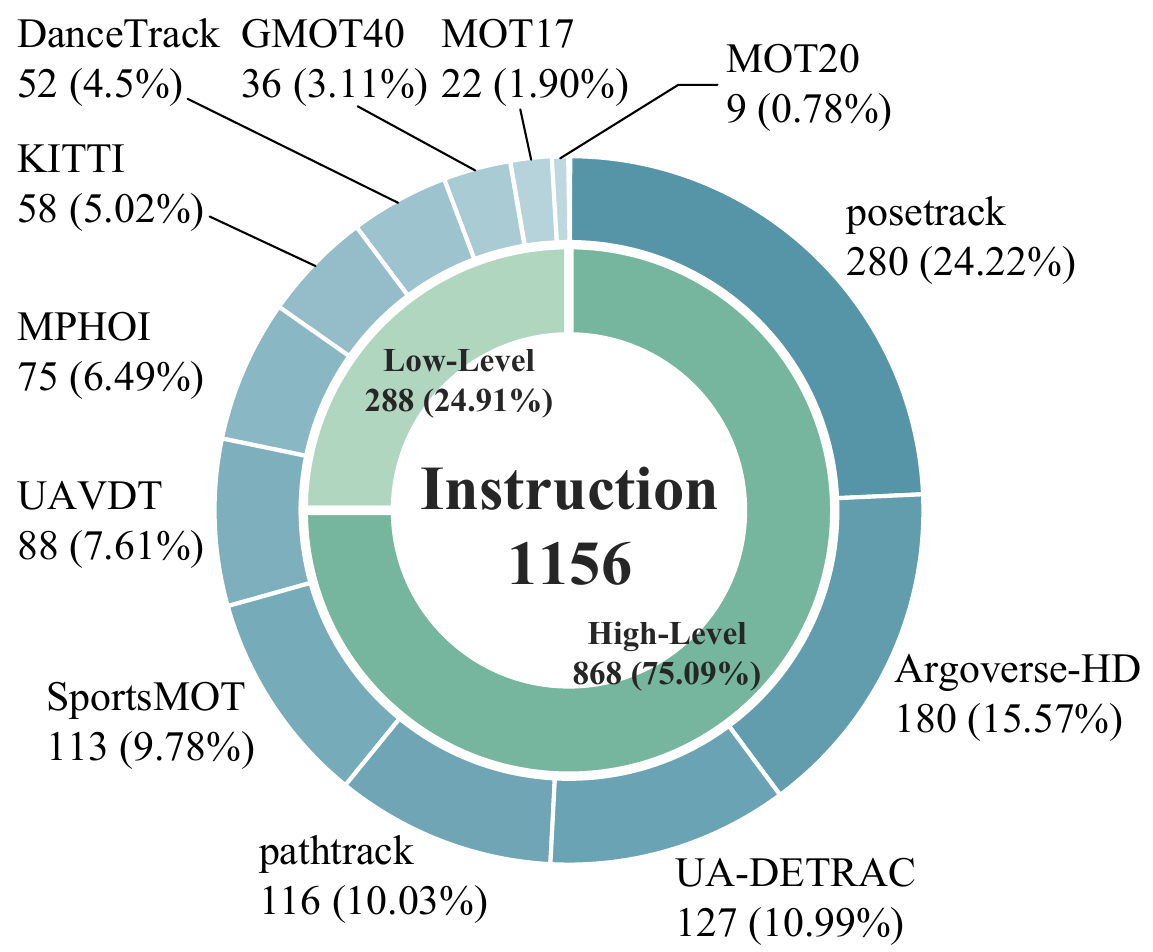}}
\hfil
\subfloat[\label{fig:d}]{
    \includegraphics[width=0.185\linewidth]{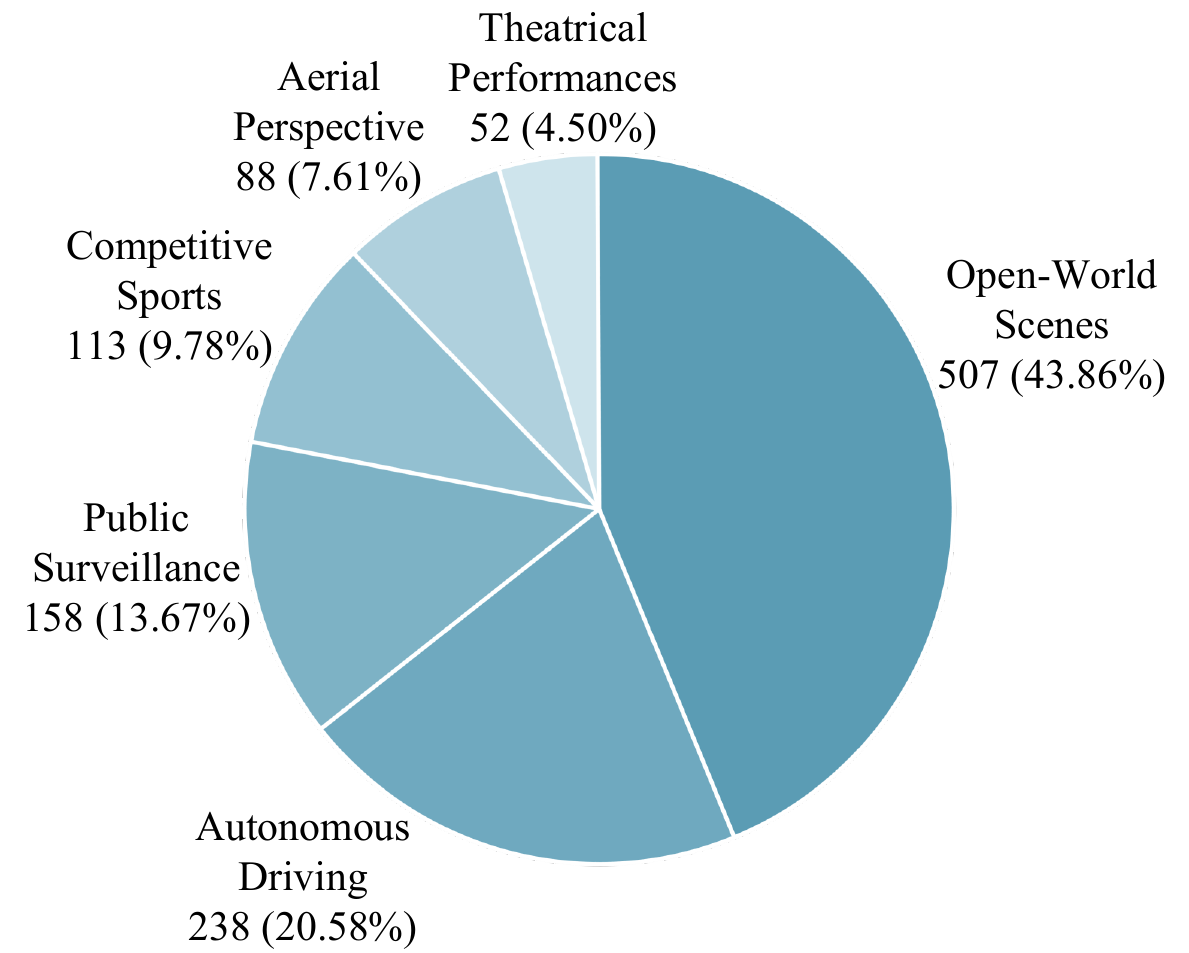}}
\caption{\textbf{Statistics of the ReaMOT Challenge dataset.} \textbf{(a) Word Cloud:} Visualizes the vocabulary distribution of language instructions, blending explicit visual cues with reasoning semantics. \textbf{(b) Instruction Sequence Length Distribution:} Displays the frame count distribution per instruction, showcasing highly diverse temporal spans that rigorously test long-term tracking stability. \textbf{(c) Instruction Level and Source Dataset Distribution:} Breaks down the language instructions by their 12 source datasets (outer ring) and reasoning levels (inner ring), highlighting the benchmark's reasoning-centric focus. \textbf{(d) Evaluation Scenario Distribution:} Categorizes the dataset into six distinct domains, showcasing comprehensive coverage of unconstrained real-world environments.}
\label{fig:Statistics of the ReaMOT Challenge dataset.}
\end{figure*}

\subsubsection{Reasoning level classification}
\label{subsubsec:Reasoning level classification}

We classify all language instructions into two levels based on the depth of reasoning required: High-Level Reasoning and Low-Level Perception.

(1) \textbf{High-Level Reasoning:} This level entails semantic reasoning derived from visual cues, requiring the integration of social common sense, logical relations, or cultural context. Instructions are classified into this level when they demand judgments regarding functional spaces, inferences about social labels and attributes, causal logic deductions, or predictions of psychological intentions.

(2) \textbf{Low-Level Perception:} This level describes direct perception of the physical world and involves only objective facts that can be obtained through basic human sensory perception, such as an object’s shape, color, spatial position, and basic body movements. This level follows a “what you see is what you get” principle and does not involve any inference about causes, rules, or mental states.

Overall, Low-Level Perception corresponds to pure physical observation, whereas High-Level Reasoning reflects deeper semantic understanding that integrates human social knowledge. Typical examples are presented in Table~\ref{tab:Core Attribute Domains}.

\subsubsection{Dataset Split}
\label{subsubsec:Dataset Split}

Of the 1,156 total instructions, we employ a random stratified sampling strategy to partition the dataset into a training set of 956 instructions and a test set of 200 instructions. Specifically, the test set comprises 140 instructions categorized as High-Level Reasoning and 60 classified as Low-Level Perception. This specific distribution places a greater emphasis on High-Level Reasoning to rigorously examine the complex reasoning capabilities of the model while still verifying its foundational perceptual performance. Furthermore, to ensure methodological rigor, we adopt a comprehensive cross-coverage strategy. This means that the instructions within each reasoning level independently span across all 12 source datasets. By ensuring that both the Low-Level and High-Level subsets cover diverse scene distributions, we avoid scene bias in any specific difficulty category. This data split strategy allows for a systematic assessment of the model’s ability to generalize from basic perceptual understanding to complex logical reasoning across various scenarios.

\subsubsection{Dataset Statistics} 
The constructed ReaMOT Challenge dataset encompasses 1,156 language instructions (868 for High-Level Reasoning and 288 for Low-Level Perception). These instructions span 869 distinct video sequences and cover 21 diverse target categories, yielding a total of 423,359 image-language pairs. A comprehensive statistical analysis of the dataset is detailed below and visualized in Fig.~\ref{fig:Statistics of the ReaMOT Challenge dataset.}.

(1) \textbf{Word Cloud.} As illustrated in Fig.~\ref{fig:Statistics of the ReaMOT Challenge dataset.} (a), the vocabulary distribution inherently reflects our dual-level reasoning hierarchy. On one hand, prominent nouns and verbs (e.g., “people”, “vehicle”, “standing”) capture explicit visual cues corresponding to Low-Level Perception. On the other hand, the dataset is heavily characterized by a rich, reasoning-driven lexicon. The prevalence of terms denoting social relationships (e.g., “team”, “player”, “couple”), underlying intents (e.g., “waiting”, “trying”, “need”), and probabilistic uncertainty (e.g., “likely”, “might”, “may”) underscores the cognitive complexity of the task. This linguistic depth ensures that ReaMOT transcends basic visual recognition, demanding deep semantic comprehension to decode implicit relationships.

(2) \textbf{Instruction Sequence Length Distribution.} Fig.~\ref{fig:Statistics of the ReaMOT Challenge dataset.} (b) visualizes the distribution of frame counts per language instruction. Across all sequence length intervals, High-Level Reasoning instructions consistently dominate, validating our core objective of building a reasoning-centric benchmark. Furthermore, the temporal span of these instructions exhibits remarkable diversity, ranging from short clips of under 200 frames to exceptionally long sequences exceeding 1,000 frames. This pronounced variance in sequence length introduces critical challenges for temporal association, rigorously testing the trackers' ability to maintain identity consistency and resist drift over extended periods.

(3) \textbf{Instruction Level and Source Dataset Distribution.} While the 12 source datasets were originally curated for conventional MOT tasks without language annotations, we have meticulously equipped their video sequences with language instructions. Fig.~\ref{fig:Statistics of the ReaMOT Challenge dataset.} (c) illustrates the distribution of these newly annotated instructions across the diverse video sources and their corresponding reasoning levels. The nested distribution highlights that PoseTrack (24.22\%), alongside autonomous driving datasets like Argoverse-HD (15.57\%) and UA-DETRAC (10.99\%), constitute the largest proportions of the source data. This composition provides a highly demanding visual foundation characterized by comprehensive scenario coverage. Concurrently, the inner ring corroborates the reasoning focus of our annotation efforts, with High-Level instructions accounting for the vast majority (75.09\%) of the benchmark compared to Low-Level Perception.

(4) \textbf{Evaluation Scenario Distribution.} To comprehensively assess model generalization, we categorize the dataset into six distinct evaluation scenarios, as shown in Fig.~\ref{fig:Statistics of the ReaMOT Challenge dataset.} (d). “Open-World Scenes” forms the largest subset (43.86\%), ensuring models are tested against unconstrained real-world environments. This is complemented by highly structured and challenging domains, including “Autonomous Driving” (20.58\%), “Public Surveillance” (13.67\%), and “Competitive Sports” (9.78\%). This diverse multi-scenario distribution ensures that the ReaMOT Challenge effectively evaluates tracking robustness across varying camera perspectives, target densities, and dynamic movement patterns.

\begin{table}[t]
\setlength{\abovecaptionskip}{0.6mm}
\centering

\caption{\textbf{Evaluation metrics for the ReaMOT task.} Metrics are averaged over $n$ language instructions, indexed by $i$. Computations utilize localization thresholds $\mathcal{A}$, Detection Accuracy $\text{DetA}$, and Association Accuracy $\text{AssA}$. Temporal identity relies on Identity True Positives ($\text{IDTP}$), False Positives ($\text{IDFP}$), and False Negatives ($\text{IDFN}$). Frame level assessment incorporates total ground truth targets ($\text{GT}$), True Positives ($\text{TP}$), False Positives ($\text{FP}$), False Negatives ($\text{FN}$), and Identity Switches ($\text{IDSW}$).}
\label{tab:Evaluation metrics of the ReaMOT task.}
\resizebox{1.0 \linewidth}{!}{
    \begin{tabular}{l|l|l}
    \toprule
    \rowcolor[HTML]{BFDAE9}
    \textbf{Metric} & \textbf{Measure Ability} & \textbf{Definition} \\
    \midrule
    RHOTA & Holistic Reasoning and Tracking & $\text{RHOTA} = \frac{1}{n} \sum_{i=1}^{n} \left( \frac{1}{|\mathcal{A}|} \sum_{\alpha \in \mathcal{A}} \sqrt{\text{DetA}_{i,\alpha} \cdot \text{AssA}_{i,\alpha}} \right)$ \\
    \midrule
    RIDF1 & Temporal Association Stability & $\text{RIDF1} = \frac{1}{n} \sum_{i=1}^{n} \frac{2 \cdot \text{IDTP}_i}{2 \cdot \text{IDTP}_i + \text{IDFP}_i + \text{IDFN}_i}$ \\
    \midrule
    RMOTA & Trajectory Integrity & $\text{RMOTA} = \frac{1}{n} \sum_{i=1}^{n} \max \left( 1 - \frac{\text{FN}_i + \text{FP}_i + \text{IDSW}_i}{\text{GT}_i}, 0 \right)$ \\
    \midrule
    RRcll & Reasoning Completeness & $\text{RRcll}=\frac{1}{n} \sum_{i=1}^n \frac{\text{TP}_i}{\text{TP}_i+\text{FN}_i}$ \\
    \midrule
    RPrcn & Reasoning Discrimination & $\text{RPrcn}=\frac{1}{n} \sum_{i=1}^n \frac{\text{TP}_i}{\text{TP}_i+\text{FP}_i}$ \\
    \bottomrule
    \end{tabular}
}
\end{table}

\subsection{Metrics Design}
\label{subsec:Metrics Design}

To evaluate reasoning capability and tracking robustness in the ReaMOT task, we employ five metrics: Higher Order Tracking Accuracy (HOTA), Identification F1 Score (IDF1), Multi Object Tracking Accuracy (MOTA), Recall (Rcll), and Precision (Prcn). Unlike generic MOT, ReaMOT scenarios feature numerous distractors visually similar to described targets. Consequently, reasoning failures can generate excessive false positives, driving standard MOTA to large negative values. To prevent this, we bound the metric within $[0, 1]$ using $\max(\text{MOTA}, 0)$. Furthermore, to ensure balanced evaluation across diverse language instructions, we compute final results by averaging scores per language instruction. This formulation yields our tailored metrics: RHOTA, RIDF1, RMOTA, RRcll, and RPrcn, as detailed in Table~\ref{tab:Evaluation metrics of the ReaMOT task.}.

\begin{figure*}[t]
\setlength{\abovecaptionskip}{1.2mm}
\centering
    \includegraphics[width=1.0\linewidth]{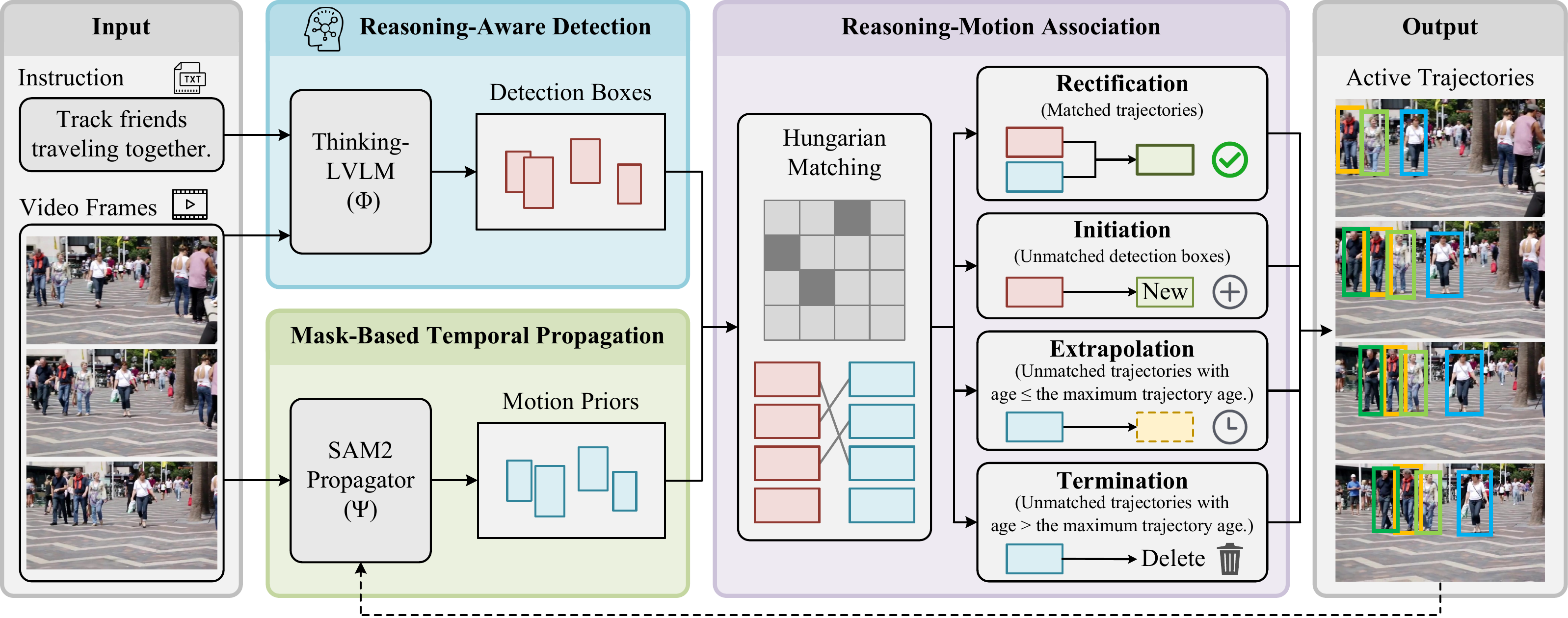}
    \caption{\textbf{Overall pipeline of ReaTrack.} The framework consists of three synergistic modules: \textbf{(a) Reasoning-Aware Detection}, which employs a Thinking-LVLM to interpret complex implicit instructions and generate semantic detections; \textbf{(b) Mask-Based Temporal Propagation}, which leverages SAM2 to predict robust physical motion priors based on historical states; and \textbf{(c) Reasoning-Motion Association}, a core decision unit that optimally aligns these semantic detections with temporal predictions to manage trajectory lifecycles (Rectification, Initiation, Extrapolation, and Termination).}
    \label{fig:Overall pipeline of ReaTrack.}
\end{figure*}

\section{Methodology}
\label{sec:Methodology}

As introduced in Section~\ref{sec:Benchmark}, we construct the ReaMOT Challenge dataset to push the boundaries of tracking from basic perception to complex cognitive reasoning. Building upon the analysis of existing works \cite{wu2023referring, zhang2024bootstrapping, du2024ikun, chen2025cross, li2025lamot, li2025language, liang2025cognitive}, it becomes evident that current RMOT paradigms fundamentally struggle with this leap. The reason is twofold: (1) traditional tracking architectures rely on explicit visual-textual matching, lacking the cognitive capacity to decode implicit logical deductions; and (2) while Large Vision-Language Models (LVLMs) \cite{liu2024visual, chen2024internvl, bai2025qwen2, bai2025qwen3} exhibit profound reasoning abilities, directly applying them for frame-by-frame tracking yields severe temporal inconsistencies and trajectory fragmentation, as they inherently lack video-level motion priors.

Driven by this insight, we argue that an effective reasoning-based tracker must organically decouple high-level cognitive localization from low-level physical motion continuity. To this end, we propose \textbf{ReaTrack}, a strong training-free baseline framework explicitly designed to bridge this gap.

\subsection{Overview}
As illustrated in Fig.~\ref{fig:Overall pipeline of ReaTrack.}, ReaTrack comprises three synergistic components, each designed to address the specific bottlenecks identified in our analysis:

\begin{itemize}
    \item \textbf{Reasoning-Aware Detection (RAD):} This module addresses the cognitive deficit of traditional methods. By leveraging a Thinking-variant LVLM, it decodes complex implicit instructions to provide highly reliable semantic detections at the frame level.
    \item \textbf{Mask-Based Temporal Propagation (MBTP):} This module overcomes the temporal inconsistency inherent to LVLMs. It utilizes SAM2's robust video memory to propagate physical object states smoothly across frames, ensuring trajectory continuity.
    \item \textbf{Reasoning-Motion Association (RMA):} Serving as the core decision-making unit, this module dynamically aligns the semantic detections from RAD with the physical motion priors from MBTP. It uses high-level reasoning to rectify accumulated spatial drift while relying on low-level motion predictions to handle short-term reasoning failures.
\end{itemize}

\subsection{Reasoning-Aware Detection}
To accurately locate objects described by complex instructions, we employ the Thinking-variant Large Vision-Language Model (LVLM) \cite{bai2025qwen3}, denoted as $\Phi$. At frame $t$, given the image $I_t \in \mathbb{R}^{H \times W \times 3}$ and the language instruction $L$, the LVLM decodes the implicit semantic cues to generate a set of discrete detections:
\begin{equation}
    \mathcal{D}_{t}^{det} = \Phi(I_t, L) = \left\{ d_i = b_{i}^{det} \right\}_{i=1}^{M_t}
\end{equation}
where $\mathcal{D}_{t}^{det}$ denotes the set of $M_t$ detected candidates at the current frame, and $b_{i}^{det} \in \mathbb{R}^4$ represents the spatial bounding box coordinates of the $i$-th detection.

\subsection{Mask-Based Temporal Propagation}
While LVLMs provide high semantic accuracy, they process frames independently, causing severe temporal inconsistency in video sequences. To address this, we introduce the Segment Anything Model 2 (SAM2) \cite{ravi2024sam}, denoted as $\Psi$, as a robust mask propagator. Building upon the promptable segmentation paradigm of the original SAM \cite{kirillov2023segment}, SAM2 extends this capability to the video domain by incorporating a spatio-temporal memory bank and memory attention mechanisms.

Specifically, let $\mathcal{T}_{t-1} = \{\mathcal{T}_k\}_{k=1}^{K}$ be the set of active trajectories at frame $t-1$. For each trajectory $\mathcal{T}_k$, we utilize its previous definitive bounding box $b_{t-1}^k$ to construct a visual prompt $\mathcal{P}_{t}^k = \{b_{t-1}^k\}$. The temporal propagation to the current frame $t$ is formulated as:
\begin{equation}
    \mathcal{D}_{t}^{pred} = \left\{ p_{k} \!=\! b_{t}^{pred, k} \mid (m_{t}^{k}, b_{t}^{pred, k}) \!=\! \Psi(I_t, \mathcal{P}_{t}^k) \right\}_{k=1}^{K}
\end{equation}
where $\mathcal{D}_{t}^{pred}$ represents the set of all predicted temporal priors, $m_{t}^{k}$ is the predicted segmentation mask for the $k$-th trajectory, and $b_{t}^{pred, k}$ is the derived bounding box.

By leveraging the continuous spatio-temporal memory, this propagation mechanism enables our framework to robustly maintain physical states through complex object dynamics, including non-linear motion and short-term occlusions.

\subsection{Reasoning-Motion Association}

Algorithm~\ref{alg:Reasoning-Motion Association} outlines the detailed procedure of our unified Reasoning-Motion Association (RMA) module. This module serves as the core decision-making unit, formally aligning the semantic detections $\mathcal{D}_t^{det}$ from the Thinking-LVLM with the temporal motion priors $\mathcal{D}_t^{pred}$ propagated by SAM2.

(1) \textbf{Optimization Formulation.} We formulate the association as a bipartite graph matching problem. The cost matrix $C \in \mathbb{R}^{M_t \times K}$ quantifies the spatial discrepancy between cognitive detections and physical predictions:

\begin{algorithm}[H]
\caption{\textbf{Reasoning-Motion Association}}
\label{alg:Reasoning-Motion Association}
\textbf{Input}: Reasoning detections ($\mathcal{D}_{t}^{det}$); 
  motion priors ($\mathcal{D}_{t}^{pred}$) \\
\textbf{Parameter}: Matched trajectories ($\mathcal{T}^{m}$); unmatched trajectories ($\mathcal{T}^{u}$); matched detection boxes ($\mathcal{D}^{m}$); unmatched detection boxes ($\mathcal{D}^{u}$); the age of trajectory $\mathcal{T}_{k}$ ($\mathcal{A}_{k}$); the maximum trajectory age (${A}_{max}$) \\
\textbf{Output}: Active trajectories ($\mathcal{T}^{active}$)

\begin{algorithmic}[1]
    \STATE Let $\mathcal{T}^{active} \leftarrow \emptyset$.
    \STATE /* Compute Cost Matrix */
    \FOR{each $d_i \in \mathcal{D}_{t}^{det}$ and $p_k \in  \mathcal{D}_{t}^{pred}$}
        \STATE $C_{i,k} \leftarrow 1 - \text{IoU}(d_i, p_k)$
    \ENDFOR
    \STATE /* Hungarian Matching */
    \STATE $\mathcal{T}^{m}, \mathcal{T}^{u}, \mathcal{D}^{m}, \mathcal{D}^{u} \leftarrow \text{Hungarian}(C)$
    \STATE /* Rectification */
    \FOR{each matched pair $(d_{i}, \mathcal{T}_{k}) \in (\mathcal{D}^{m}, \mathcal{T}^{m}$)}
        \STATE Update state of $\mathcal{T}_k$ using matched detection $d_{i}$
        \STATE $\mathcal{A}_{k} \leftarrow 0$
        \STATE $\mathcal{T}^{active} \leftarrow \mathcal{T}^{active} \cup \{\mathcal{T}_{k}\}$
    \ENDFOR
    \STATE /* Initiation */
    \FOR{each $d_{j} \in \mathcal{D}^{u}$}
        \STATE Initialize new track $\mathcal{T}^{new}$ from $d_j$
        \STATE $\mathcal{T}^{active} \leftarrow \mathcal{T}^{active} \cup \{\mathcal{T}^{new}\}$
    \ENDFOR
    \STATE /* Extrapolation \& Termination */ 
    \FOR{each $\mathcal{T}_{k} \in \mathcal{T}^{u}$}
        \IF {$\mathcal{A}_{k} \leq {A}_{max}$}
            \STATE Update state of $\mathcal{T}_k$ using motion prior $p_k$
            \STATE $\mathcal{A}_{k} = \mathcal{A}_{k} + 1$            
            \STATE $\mathcal{T}^{active} \leftarrow \mathcal{T}^{active} \cup \{\mathcal{T}_{k}\}$
        \ENDIF
    \ENDFOR
    \STATE \textbf{return} $\mathcal{T}^{active}$
\end{algorithmic}
\end{algorithm}

\begin{equation}
    C_{i,k} = 1 - \text{IoU}\left(b_{i}^{det}, b_{t}^{pred, k}\right)
\end{equation}
where $C_{i,k}$ is the matching cost between the $i$-th semantic detection and the $k$-th motion prior. 

The optimal assignment matrix $X^* \in \{0, 1\}^{M_t \times K}$ is then obtained by solving the objective function via the Hungarian algorithm \cite{kuhn1955hungarian}:
\begin{equation}
    X^* = \arg\min_{X} \sum_{i=1}^{M_t} \sum_{k=1}^{K} C_{i,k} X_{i,k}
\end{equation}
where binary variable $X_{i,k} \in \{0, 1\}$ indicates whether the $i$-th detection is matched to the $k$-th trajectory. This optimization is subject to the constraints $\sum_{i} X_{i,k} \leq 1$ and $\sum_{k} X_{i,k} \leq 1$, ensuring a mutually exclusive one to one matching. The matrix $X^*$ partitions the inputs into matched pairs ($\mathcal{T}^{m}, \mathcal{D}^{m}$), unmatched detections ($\mathcal{D}^{u}$), and unmatched trajectories ($\mathcal{T}^{u}$).

(2) \textbf{State Evolution Logic.} Based on the optimal matching matrix $X^*$, the spatial state $S_{t}^{k}$ for trajectory $\mathcal{T}_k$ at frame $t$ is updated as follows:
\begin{equation}
    S_{t}^{k} = 
    \begin{cases}
        b_{i}^{det}, & \text{if } X_{i,k}^* = 1 \\
        b_{t}^{pred, k}, & \text{if } \sum_{i} X_{i,k}^* = 0 \text{ and } \mathcal{A}_k \leq A_{max} \\
        \emptyset, & \text{otherwise}
    \end{cases}
\end{equation}
where $S_{t}^{k}$ denotes the definitive spatial state of trajectory $\mathcal{T}_k$ at frame $t$, $\mathcal{A}_k$ represents the trajectory age (number of consecutive unmatched frames), and $A_{max}$ is the predefined maximum trajectory age threshold.

\textbf{Rectification:} For matched trajectories, the definitive state $S_{t}^{k}$ is updated by the LVLM semantic detection $b_{i}^{det}$. This design leverages the reasoning reliability of the LVLM to precisely correct the trajectory, eliminating any cumulative spatial drift generated by the mask propagator. The trajectory age $\mathcal{A}_k$ is reset to 0.

\textbf{Initiation:} Unmatched semantic detections ($\mathcal{D}^u$) are treated as definitive semantic proposals. We initialize new independent trajectory instances for them to capture newly appearing or semantically re-entering targets.

\textbf{Extrapolation \& Termination:} Unmatched trajectories represent tracking failures due to either physical occlusion or LVLM reasoning dropouts. To guarantee trajectory integrity, we execute a “coasting” extrapolation, temporarily adopting the SAM2 motion prior $b_{t}^{pred, k}$ as the definitive state $S_{t}^{k}$ while incrementing $\mathcal{A}_k$. If the trajectory remains undetected beyond the threshold $A_{max}$, it is permanently terminated ($S_{t}^{k} = \emptyset$) to suppress long-term false positives.

This mathematically decoupled logic guarantees that ReaTrack robustly maintains temporal continuity during intermittent reasoning failures, while instantaneously rectifying physical drift once a concrete cognitive signal re-emerges.

\section{Experiments}
\label{sec:Experiments}

\subsection{Implementation Details}
\label{subsec:Implementation Details}

We implement ReaTrack as a training-free framework utilizing \textit{Qwen3-VL-8B-Thinking} for reasoning and \textit{sam2\_hiera\_large} for mask propagation. For LVLM inference, we set \textit{max\_new\_tokens} to 1024 for the \textit{Thinking} model to accommodate comprehensive reasoning chains, and 128 for the \textit{Instruct} variant. The maximum trajectory age ${A}_{max}$ is set to 10 frames. All experiments are conducted on a single NVIDIA RTX A6000 GPU.

To accommodate the specific input formatting requirements of various Large Vision Language Models (LVLMs) evaluated in our study, we tailor the instruction prompts, dynamically inserting the target description as \textit{$\langle$language instruction$\rangle$}. For our primary models, \textit{Qwen3-VL-8B-Thinking} and \textit{Qwen3-VL-8B-Instruct}, the prompt is formulated as: “Please detect and label all \textit{$\langle$language instruction$\rangle$} in the following image and mark their positions. Return the 2D bounding boxes in image pixel coordinates.” (e.g., \textit{“Please detect and label all the attacking players in the following image and mark their positions. Return the 2D bounding boxes in image pixel coordinates.”}). For \textit{DeepSeek-VL-7B-chat}, \textit{LLaVA-1.5-7B}, and \textit{LLaVA-NEXT-8B}, we input: “Please detect all \textit{$\langle$language instruction$\rangle$} in the image and output their coordinates with [x1, y1, x2, y2] format.” Finally, for \textit{InternVL3.5-8B}, the prompt is designed as: “Please detect and label all \textit{$\langle$language instruction$\rangle$} in the following image and mark their positions.”

\begin{table*}[t]
\centering
\setlength{\abovecaptionskip}{0.6mm}
    \caption{\textbf{Overall quantitative performance} comparison of state-of-the-art methods on the ReaMOT Challenge benchmark under zero-shot settings. ↑ indicates that higher score is better. The best results are marked in \textbf{bold}.}
    \label{tab:Overall quantitative performance.}
    \resizebox{1.0 \linewidth}{!}{
        \setlength{\tabcolsep}{1.5mm}{
        \begin{tabular}{lc ccccc ccccc}
            \toprule
            \multirow{3}{*}{\textbf{Method}} & \multirow{3}{*}{\textbf{Published}} & \multicolumn{5}{c}{\cellcolor{Color-High-Level}\textbf{High-Level Reasoning}} & \multicolumn{5}{c}{\cellcolor{Color-Low-Level}\textbf{{Low-Level Perception}}} \\
            \cmidrule(lr){3-7} \cmidrule(lr){8-12}
            \rule{0pt}{10pt} & & \textbf{RHOTA↑} & \textbf{RIDF1↑} & \textbf{RMOTA↑} & \textbf{RRcll↑} & \textbf{RPrcn↑} & \textbf{RHOTA↑} & \textbf{RIDF1↑} & \textbf{RMOTA↑} & \textbf{RRcll↑} & \textbf{RPrcn↑} \\
            \midrule
            \multicolumn{2}{l}{\textbf{End-to-End Methods:}} & \multicolumn{10}{l}{} \\
            \midrule
            TransRMOT \cite{wu2023referring} & CVPR 2023 & 1.97 & 0.65 & 0.13 & 0.51 & 9.34 & 6.04 & 2.87 & 1.02 & 2.68 & 25.31  \\
            TempRMOT \cite{zhang2024bootstrapping} & arXiv 2024 & 6.58 & 4.71 & 2.14 & 3.79 & 14.85 & 12.95 & 8.27 & 2.83 & 6.47 & 23.55  \\
            CRTracker \cite{chen2025cross} & AAAI 2025 & 11.91 & 8.54 & 2.44 & 9.90 & 20.94 & 9.03 & 7.76 & 4.77 & 7.89 & 19.01 \\
            DKGTrack \cite{li2025language} & ICCV 2025 & 2.80 & 1.41 & 0.46 & 0.94 & 8.90 & 6.94 & 4.17 & 1.93 & 2.75 & 20.58 \\
            \midrule
            \multicolumn{2}{l}{\textbf{Two-Stage Methods:}} & \multicolumn{10}{l}{} \\
            \midrule
            YOLOX+ByteTrack+iKUN \cite{du2024ikun} & CVPR 2024 & 12.66 & 9.54 & 1.44 & 17.94 & 9.87 & 7.59 & 5.90 & 1.71 & 7.77 & 8.72 \\
            YOLOX+OC-SORT+iKUN \cite{du2024ikun} & CVPR 2024 & 13.00 & 10.00 & 1.83 & 17.70 & 10.42 & 7.07 & 5.08 & 1.13 & 6.74 & 7.67 \\
            \midrule
            \rowcolor[HTML]{E6E6E6} \textbf{ReaTrack (Ours)} & - & \textbf{42.28} & \textbf{38.58} & \textbf{16.30} & \textbf{55.13} & \textbf{37.83} & \textbf{31.55} & \textbf{27.58} & \textbf{8.08} & \textbf{40.43} & \textbf{27.41} \\
            \bottomrule
        \end{tabular}
        }
    }
\vspace{-4pt}
\end{table*}

\subsection{Comparison with State-of-the-Art Methods}

To comprehensively evaluate ReaTrack, we conduct extensive comparisons against state-of-the-art methods on the ReaMOT Challenge benchmark under a zero-shot setting. Our evaluation encompasses two primary tracking paradigms:

(1) \textbf{End-to-End Methods:} We select four representative open-source RMOT models: TransRMOT \cite{wu2023referring}, TempRMOT \cite{zhang2024bootstrapping}, CRTracker \cite{chen2025cross}, and DKGTrack \cite{li2025language}. 

(2) \textbf{Two-Stage Methods:} We construct a robust two-stage pipeline using the latest open-source RMOT algorithm, iKUN \cite{du2024ikun}. To handle the extensive category diversity in the ReaMOT Challenge dataset, we first employ a YOLOX-x detector \cite{ge2021yolox} (pre-trained on the 80-category COCO2017 dataset \cite{lin2014microsoft}) to extract frame-level bounding boxes. These detections are then linked into reliable trajectories by established MOT trackers, specifically ByteTrack \cite{zhang2022bytetrack} and OC-SORT \cite{cao2023observation}. Finally, iKUN computes vision-language similarity scores for these trajectories, filtering out those below a predefined threshold to yield the final predictions.

\textbf{Overall Performance Analysis.} The overall quantitative results are summarized in Table~\ref{tab:Overall quantitative performance.}. On the High-Level Reasoning subset, ReaTrack establishes a new state-of-the-art, significantly outperforming all competing models. Most notably, it achieves an impressive RHOTA of 42.28\%, surpassing the best-performing two-stage baseline (YOLOX + OC-SORT + iKUN at 13.00\%) by a massive absolute margin of 29.28\%. Similar overwhelming gains are observed in RIDF1 (38.58\%)

\begin{table*}[t]
\centering
\setlength{\abovecaptionskip}{0.6mm}
    \caption{\textbf{RHOTA performance} comparison of state-of-the-art methods across \textbf{six diverse evaluation scenarios} on the ReaMOT Challenge benchmark under zero shot settings. \textbf{HL} and \textbf{LL} denote \textbf{High Level Reasoning} and \textbf{Low Level Perception}, respectively. The best results are marked in \textbf{bold}.}
    \label{tab:scenario performance.}
    \resizebox{1.0 \linewidth}{!}{
        \setlength{\tabcolsep}{1.5mm}{
        \begin{tabular}{lc cc cc cc cc cc cc}
            \toprule
            \multirow{5}{*}{\textbf{Method}} & \multirow{5}{*}{\textbf{Published}} & \multicolumn{12}{c}{\textbf{Evaluation Scenarios}} \\
            \cmidrule(lr){3-14}
            & & \multicolumn{2}{c}{\textbf{Open-World}} & \multicolumn{2}{c}{\textbf{Autonomous}} & \multicolumn{2}{c}{\textbf{Public}} & \multicolumn{2}{c}{\textbf{Competitive}} & \multicolumn{2}{c}{\textbf{Aerial}} & \multicolumn{2}{c}{\textbf{Theatrical}} \\
            & & \multicolumn{2}{c}{\textbf{Scenes}} & \multicolumn{2}{c}{\textbf{Driving}} & \multicolumn{2}{c}{\textbf{Surveillance}} & \multicolumn{2}{c}{\textbf{Sports}} & \multicolumn{2}{c}{\textbf{Perspective}} & \multicolumn{2}{c}{\textbf{Performances}} \\
            \cmidrule(lr){3-4} \cmidrule(lr){5-6} \cmidrule(lr){7-8} \cmidrule(lr){9-10} \cmidrule(lr){11-12} \cmidrule(lr){13-14}
            & & \cellcolor{Color-High-Level}\textbf{HL} & \cellcolor{Color-Low-Level}\textbf{LL} & \cellcolor{Color-High-Level}\textbf{HL} & \cellcolor{Color-Low-Level}\textbf{LL} & \cellcolor{Color-High-Level}\textbf{HL} & \cellcolor{Color-Low-Level}\textbf{LL} & \cellcolor{Color-High-Level}\textbf{HL} & \cellcolor{Color-Low-Level}\textbf{LL} & \cellcolor{Color-High-Level}\textbf{HL} & \cellcolor{Color-Low-Level}\textbf{LL} & \cellcolor{Color-High-Level}\textbf{HL} & \cellcolor{Color-Low-Level}\textbf{LL} \\
            \midrule
            \multicolumn{2}{l}{\textbf{End-to-End Methods:}} & \multicolumn{6}{l}{} \\
            \midrule
            TransRMOT \cite{wu2023referring} & CVPR 2023 & 0.97 & 3.01 & 5.30 & 11.07 & 4.31 & 7.85 & 0.68 & 3.73 & 0.51 & \textbf{3.25} & 0.39 & 7.62 \\
            TempRMOT \cite{zhang2024bootstrapping} & arXiv 2024 & 8.17 & 10.34 & 13.57 & 23.57 & 2.78 & 11.53 & 1.25 & 5.22 & 1.04 & 0.55 & 0.90 & 9.33 \\
            CRTracker \cite{chen2025cross} & AAAI 2025 & 13.05 & 8.27 & 15.62 & 2.64 & 1.66 & 19.70 & 17.30 & 22.00 & 0.82 & 0 & 11.20 & 5.37 \\
            DKGTrack \cite{li2025language} & ICCV 2025 & 2.76 & 7.45 & 4.44 & 10.32 & 2.31 & 1.41 & 2.57 & 9.92 & 0 & 0 & 3.92 & 2.63 \\
            \midrule
            \multicolumn{2}{l}{\textbf{Two-Stage Methods:}} & \multicolumn{6}{l}{} \\
            \midrule
            YOLOX+ByteTrack+iKUN \cite{du2024ikun} & CVPR 2024 & 20.38 & 14.98 & 9.30 & 2.87 & 3.82 & 1.09 & 9.56 & 5.05 & 2.22 & 0 & 1.04 & 0 \\
            YOLOX+OC-SORT+iKUN \cite{du2024ikun} & CVPR 2024 & 20.89 & 13.84 & 10.31 & 2.71 & 3.89 & 1.07 & 9.12 & 5.07 & 2.32 & 0 & 0.92 & 0 \\
            \midrule
            \rowcolor[HTML]{E6E6E6} \textbf{ReaTrack (Ours)} & - & \textbf{49.67} & \textbf{32.59} & \textbf{47.14} & \textbf{38.02} & \textbf{36.69} & \textbf{25.73} & \textbf{37.05} & \textbf{36.17} & \textbf{17.87} & 0 & \textbf{27.96} & \textbf{57.49} \\
            \bottomrule
        \end{tabular}
        }
    }
\end{table*}

\noindent and RMOTA (16.30\%). This immense performance gap exposes the fundamental limitations of traditional RMOT architectures in processing complex, implicit semantic instructions. Furthermore, this architectural advantage naturally extends to the Low-Level Perception subset. Even when tasked with explicitly descriptive instructions, ReaTrack maintains a dominant lead, achieving a RHOTA of 31.55\% compared to the best baseline score of 12.95\% (TempRMOT).

\textbf{Fine-Grained Scenario Robustness Analysis.} To investigate generalization capabilities across diverse unconstrained environments, Table \ref{tab:scenario performance.} presents a detailed performance breakdown across six scenarios encompassing both High-Level Reasoning (HL) and Low-Level Perception (LL) requirements. ReaTrack demonstrates overwhelming superiority when complex reasoning is required, achieving remarkable RHOTA scores of 49.67\% and 47.14\% in highly dynamic environments such as Open-World Scenes and Autonomous Driving, respectively, whereas existing state of the art methods suffer severe degradation from visual occlusions and background clutter. Furthermore, this robustness extends to perception driven tasks, with ReaTrack vastly outperforming competitors by achieving a striking 57.49\% RHOTA in Theatrical Performances. However, an notable failure case emerges in the Aerial Perspective scenario, where ReaTrack struggles significantly. This limitation stems from extreme scale variations and minuscule target sizes inherent to drone captured footage, which directly exceed the current resolution and spatial parsing limits of the underlying LVLM visual encoders. Addressing this scale variance in reasoning driven tracking presents a highly valuable avenue for future research, while our current framework firmly establishes a robust foundation for the field.

\begin{table*}[t]
\centering
\setlength{\abovecaptionskip}{0.5mm}
    \caption{\textbf{Performance comparison of various Large Vision-Language Models (LVLMs)} in the Reasoning-Aware Detection module.}
    \label{tab:Performance comparison of various Large Vision-Language Models (LVLMs).}
    \resizebox{1.0 \linewidth}{!}{
        \centering
        \begin{tabular}{l ccccc ccccc}
            \toprule
            \multirow{3}{*}{\textbf{Reasoning-Aware Detection Module}} & \multicolumn{5}{c}{\cellcolor{Color-High-Level}\textbf{High-Level Reasoning}} & \multicolumn{5}{c}{\cellcolor{Color-Low-Level}{\textbf{Low-Level Perception}}} \\
            \cmidrule(lr){2-6} \cmidrule(lr){7-11}
            \rule{0pt}{10pt} & \textbf{RHOTA↑} & \textbf{RIDF1↑} & \textbf{RMOTA↑} & \textbf{RRcll↑} & \textbf{RPrcn↑} & \textbf{RHOTA↑} & \textbf{RIDF1↑} & \textbf{RMOTA↑} & \textbf{RRcll↑} & \textbf{RPrcn↑} \\
            \midrule
            DeepSeek-VL-7B-chat \cite{lu2024deepseek} & 8.49 & 4.36 & 0 & 28.69 & 3.06 & 7.84 & 3.62 & 0 & 26.95 & 2.45 \\
            LLaVA-1.5-7B \cite{liu2024visual} & 7.88 & 3.93 & 0.36 & 28.81 & 3.07 & 8.89 & 4.89 & 0 & 33.80 & 3.22 \\
            LLaVA-NEXT-8B \cite{liu2024llavanext} & 19.40 & 16.50 & 4.99 & 37.59 & 17.69 & 18.59 & 15.39 & 0.78 & \textbf{41.55} & 12.32 \\
            InternVL3.5-8B \cite{wang2025internvl3} & 17.10 & 12.20 & 2.02 & 18.45 & 13.77 & 12.98 & 8.21 & 1.27 & 21.66 & 9.00 \\
            \midrule
            \rowcolor[HTML]{E6E6E6} Qwen3-VL-8B-Thinking \cite{bai2025qwen3} & \textbf{42.28}  & \textbf{38.58}  & \textbf{16.30}  & \textbf{55.13}  & \textbf{37.83}  & \textbf{31.55} & \textbf{27.58} & \textbf{8.08} & 40.43  & \textbf{27.41}  \\
            \bottomrule
        \end{tabular}
    }
\vspace{-5pt}
\end{table*}

\subsection{Ablation Study}
To study the role of each part of our ReaTrack framework, we conduct ablation experiments on the ReaMOT Challenge benchmark under zero-shot settings.

\subsubsection{Comparison of Different Large Vision-Language Models (LVLMs)}
The tracking efficacy of the Reasoning-Aware Detection (RAD) module is fundamentally bounded by the cognitive capability of its underlying LVLM. As illustrated in Table~\ref{tab:Performance comparison of various Large Vision-Language Models (LVLMs).}, integrating Qwen3-VL-8B-Thinking \cite{bai2025qwen3} as the core engine yields impressive improvements across all metrics. Most strikingly, on the High-Level Reasoning subset, it achieves an RHOTA of 42.28\%, effectively more than doubling the performance of the strongest runner-up, LLaVA-NEXT-8B (19.40\%). This superiority, further evidenced by 38.58\% RIDF1 and 16.30\% RMOTA, exposes the bottleneck of conventional LVLM architectures facing implicit, logic-heavy tracking instructions. By leveraging its advanced Chain-of-Thought (\textit{Thinking}) mechanism, Qwen3-VL-8B-Thinking establishes a vastly more robust and precise mapping between the complex language instruction and the visual detection set. These results conclusively validate that a high-performing, reasoning-native engine is the indispensable foundation for resolving deep semantic ambiguities in the ReaMOT task.

\begin{table*}[t]
\centering
\setlength{\abovecaptionskip}{0.5mm}
    \caption{\textbf{Ablation study comparing the “Instruct” and “Thinking” variant} of Qwen3-VL.}
    \label{tab:Ablation study comparing the “Instruct” and “Thinking” variant.}
    \resizebox{1.0 \linewidth}{!}{
        \centering
        \begin{tabular}{l ccccc ccccc}
            \toprule
            \multirow{3}{*}{\textbf{Reasoning-Aware Detection Module}} & \multicolumn{5}{c}{\cellcolor{Color-High-Level}\textbf{High-Level Reasoning}} & \multicolumn{5}{c}{\cellcolor{Color-Low-Level}{\textbf{Low-Level Perception}}} \\
            \cmidrule(lr){2-6} \cmidrule(lr){7-11}
            \rule{0pt}{10pt} & \textbf{RHOTA↑} & \textbf{RIDF1↑} & \textbf{RMOTA↑} & \textbf{RRcll↑} & \textbf{RPrcn↑} & \textbf{RHOTA↑} & \textbf{RIDF1↑} & \textbf{RMOTA↑} & \textbf{RRcll↑} & \textbf{RPrcn↑} \\
            \midrule
            Qwen3-VL-8B-Instruct \cite{bai2025qwen3} & 38.31  & 33.90  & 12.33  & 52.46  & 35.39  & \textbf{46.83}  & \textbf{39.75}  & \textbf{17.68}  & \textbf{57.11}  & \textbf{42.59} \\
            \rowcolor[HTML]{E6E6E6} Qwen3-VL-8B-Thinking \cite{bai2025qwen3} & \textbf{42.28}  & \textbf{38.58}  & \textbf{16.30}  & \textbf{55.13}  & \textbf{37.83}  & 31.55 & 27.58 & 8.08 & 40.43  & 27.41  \\
            \bottomrule
        \end{tabular}
    }
\vspace{-5pt}
\end{table*}

\begin{table*}[t]
\centering
\setlength{\abovecaptionskip}{0.5mm}
    \caption{\textbf{Comparison between our association method (MBTP + RMA) and traditional association algorithms.}}
    \label{tab:Comparison between our association method (MBTP + RMA) and traditional association algorithms.}
    \resizebox{1.0 \linewidth}{!}{
        \centering
        \begin{tabular}{ll ccccc ccccc}
            \toprule
            \multirow{3}{*}{\textbf{Detection}} & \multirow{3}{*}{\textbf{Association}} & \multicolumn{5}{c}{\cellcolor{Color-High-Level}\textbf{High-Level Reasoning}} & \multicolumn{5}{c}{\cellcolor{Color-Low-Level}{\textbf{Low-Level Perception}}} \\
            \cmidrule(lr){3-7} \cmidrule(lr){8-12}
            \rule{0pt}{10pt} & & \textbf{RHOTA↑} & \textbf{RIDF1↑} & \textbf{RMOTA↑} & \textbf{RRcll↑} & \textbf{RPrcn↑} & \textbf{RHOTA↑} & \textbf{RIDF1↑} & \textbf{RMOTA↑} & \textbf{RRcll↑} & \textbf{RPrcn↑} \\
            \midrule
            Qwen3-VL-8B-Thinking & SORT \cite{bewley2016simple} & 33.38 & 33.39 & 25.33 & 33.74 & \textbf{63.38} & 24.31 & 22.72 & \textbf{13.61} & 25.11 & \textbf{48.74} \\
            Qwen3-VL-8B-Thinking & ByteTrack \cite{zhang2022bytetrack} & 37.92 & 35.89 & \textbf{25.66} & 33.09 & 62.45 & 28.21 & 25.66 & 13.58 & 24.38 & 48.44 \\
            Qwen3-VL-8B-Thinking & OC-SORT \cite{cao2023observation} & 34.28 & 31.73 & 23.49 & 28.42 & 61.67 & 22.45 & 19.16 & 10.02 & 18.12 & 45.29 \\
            \midrule
            \rowcolor[HTML]{E6E6E6} Qwen3-VL-8B-Thinking & MBTP + RMA & \textbf{42.28}  & \textbf{38.58}  & 16.30  & \textbf{55.13}  & 37.83  & \textbf{31.55} & \textbf{27.58} & 8.08 & \textbf{40.43}  & 27.41  \\
            \bottomrule
        \end{tabular}
    }
\vspace{-5pt}
\end{table*}

\begin{table*}[t]
\centering
\setlength{\abovecaptionskip}{0.5mm}
    \caption{\textbf{Sensitivity analysis of the maximum trajectory age $A_{max}$} on tracking performance.}
    \label{tab:Sensitivity analysis of the maximum trajectory age $A_{max}$.}
    \resizebox{1.0 \linewidth}{!}{
        \centering
        \begin{tabular}{c ccccc ccccc}
            \toprule
            \multirow{3}{*}{\textbf{Maximum trajectory age (${A}_{max}$})} & \multicolumn{5}{c}{\cellcolor{Color-High-Level}\textbf{High-Level Reasoning}} & \multicolumn{5}{c}{\cellcolor{Color-Low-Level}{\textbf{Low-Level Perception}}} \\
            \cmidrule(lr){2-6} \cmidrule(lr){7-11}
            \rule{0pt}{10pt} & \textbf{RHOTA↑} & \textbf{RIDF1↑} & \textbf{RMOTA↑} & \textbf{RRcll↑} & \textbf{RPrcn↑} & \textbf{RHOTA↑} & \textbf{RIDF1↑} & \textbf{RMOTA↑} & \textbf{RRcll↑} & \textbf{RPrcn↑} \\
            \midrule
            20 & 42.14 & 38.38 & 15.93 & \textbf{55.13} & 37.50 & \textbf{31.56} & 27.57 & \textbf{8.08} & 40.45 & \textbf{27.41} \\
            30 & 42.10 & 38.33 & 15.93 & \textbf{55.13} & 37.44 & 31.55 & 27.56 & \textbf{8.08} & 40.45 & 27.40 \\
            40 & 42.10 & 38.33 & 15.94 & \textbf{55.13} & 37.45 & \textbf{31.56} & 27.55 & \textbf{8.08} & \textbf{40.58} & 27.37 \\
            50 & 42.09 & 38.31 & 15.92 & \textbf{55.13} & 37.41 & 31.53 & 27.54 & \textbf{8.08} & \textbf{40.58} & 27.36 \\
            \midrule
            \rowcolor[HTML]{E6E6E6} 10 & \textbf{42.28} & \textbf{38.58} & \textbf{16.30} & \textbf{55.13} & \textbf{37.83} & 31.55 & \textbf{27.58} & \textbf{8.08} & 40.43 & \textbf{27.41} \\
            \bottomrule
        \end{tabular}
    }
\end{table*}

\subsubsection{Impact of Thinking vs. Instruct Models}
Table~\ref{tab:Ablation study comparing the “Instruct” and “Thinking” variant.} illustrates the critical role of the Chain-of-Thought mechanism in processing complex instructions. We observe a distinct performance inversion between the two variants depending on the task complexity. The Instruct model notably excels in Low-Level Perception tasks, achieving a dominant RHOTA of 46.83\%, as standard instruction-following architectures are highly optimized for direct and explicit visual grounding. However, it struggles significantly in scenarios requiring deep semantic deduction. Conversely, the Thinking variant substantially enhances high-level reasoning, improving the High-Level Reasoning RHOTA from 38.31\% to 42.28\% and RIDF1 from 33.90\% to 38.58\%. These results suggest that the internal reasoning steps of the Thinking model better align visual targets with multi-layered linguistic constraints, effectively mitigating the semantic shift often seen in conventional models. While this extensive reasoning process can occasionally overcomplicate simple perception instructions, resulting in lower scores on explicitly descriptive tasks, it proves essential for decoding the implicit relationships that define the core complexity of the ReaMOT task.

\subsubsection{Evaluation of Association Methods} 
The effectiveness of our proposed association paradigm is demonstrated in Table~\ref{tab:Comparison between our association method (MBTP + RMA) and traditional association algorithms.}. Using Qwen3-VL-8B-Thinking as the unified detection baseline, our MBTP+RMA strategy achieves the highest RHOTA (42.28\%) and RIDF1 (38.58\%) on the High-Level Reasoning subset, maintaining a consistent lead in Low-Level Perception as well. While our method exhibits lower RMOTA and RPrcn compared to traditional spatial-only trackers like ByteTrack, this variance reflects a deliberate and advantageous architectural trade-off. Traditional algorithms strictly optimize for frame-level precision, rendering them excessively conservative and vulnerable to prematurely terminating trajectories during periods of complex reasoning dropout. In stark contrast, our framework inherently prioritizes semantic recall and long-term trajectory integrity. As evidenced by the massive surge in RRcll (from 33.09\% to 55.13\% compared to ByteTrack), synergizing MBTP's motion priors with RMA's rectification logic allows the system to robustly bridge detection gaps and preserve target identities despite intermittent LVLM reasoning failures. Ultimately, sacrificing marginal frame-level precision secures an optimal balance between cognitive deduction and temporal continuity, driving the superior overall performance in the primary comprehensive metric, RHOTA.

\subsubsection{Sensitivity to Maximum Trajectory Age ($A_{max}$)}
Table~\ref{tab:Sensitivity analysis of the maximum trajectory age $A_{max}$.} evaluates the impact of the $A_{max}$ parameter, which governs the duration for retaining unmatched tracks during the trajectory extrapolation phase. The quantitative results demonstrate that our framework exhibits exceptional hyperparameter robustness, maintaining highly stable performance across a broad spectrum of values. Specifically, setting $A_{max} = 10$ yields the optimal configuration for High-Level Reasoning, achieving a peak RHOTA of 42.28\%. Maintaining a relatively lower threshold effectively prevents the system from retaining stale trajectories lacking semantic relevance, thereby preserving high tracking precision. This behavior highlights a core advantage of our architecture. Because the RAD module continuously performs active reasoning-based detection in every frame, the system does not need to heavily rely on long-term motion extrapolation. Consequently, avoiding excessively large $A_{max}$ values mitigates the risk of introducing false positives and identity switches that typically occur when a target is no longer actively supported by the underlying reasoning signal.

\begin{figure}[t]
\centering
\setlength{\abovecaptionskip}{0.6mm}
    \includegraphics[width=1.0\linewidth]{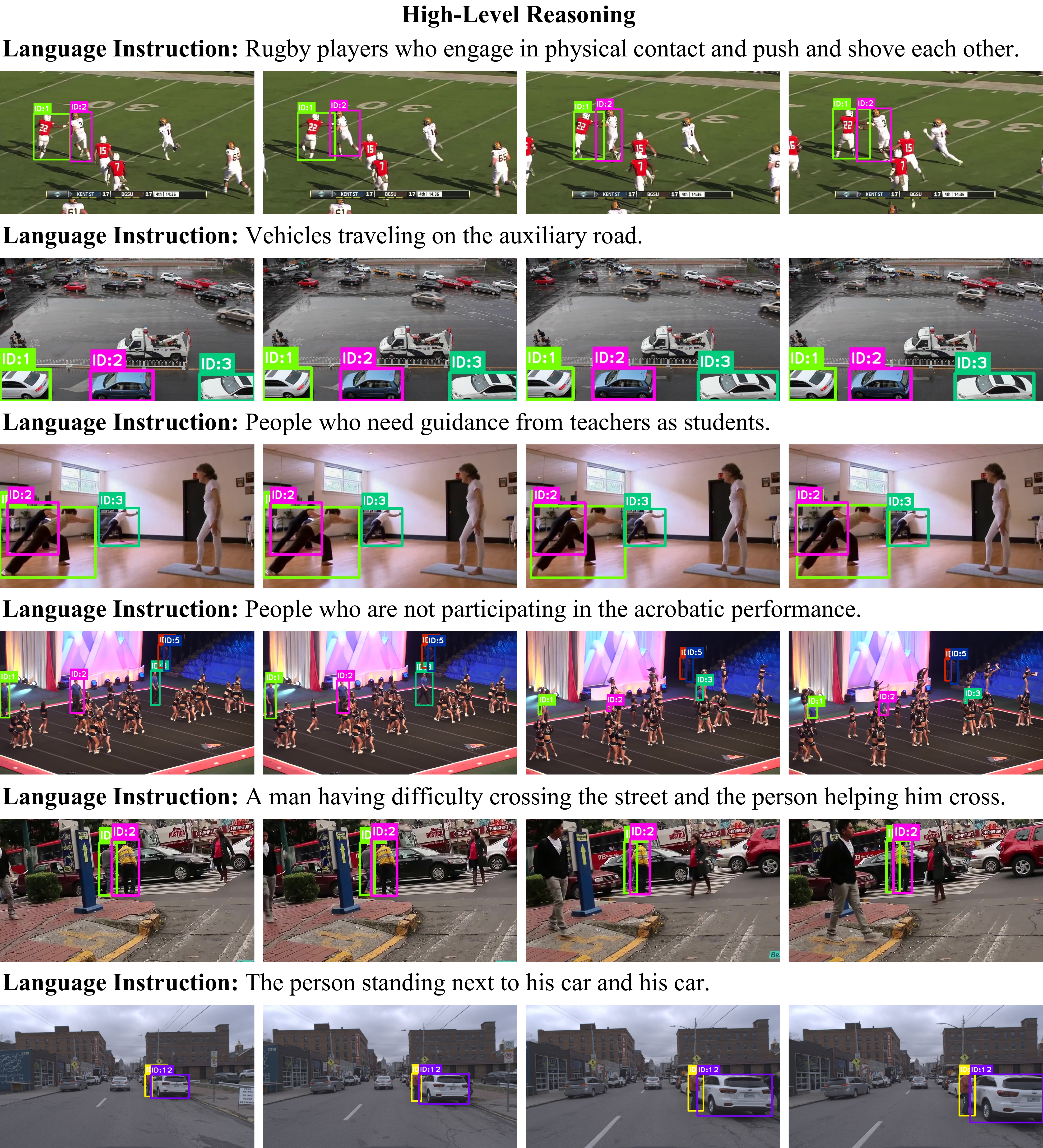}
    \caption{\textbf{Qualitative results} of the ReaTrack framework on the ReaMOT Challenge benchmark under zero-shot settings.}
    \label{fig:Qualitative Results.}
\vspace{-2pt}
\end{figure}

\subsection{Qualitative Results of the ReaTrack Framework}
\label{subsec:Appendix More Qualitative Results of the ReaTrack Framework}

As illustrated in Fig.~\ref{fig:Qualitative Results.}, qualitative evaluations under zero shot settings demonstrate ReaTrack's proficiency in synthesizing visual cues, temporal dynamics, and common sense knowledge to infer latent relationships and disambiguate distractors. The framework effectively grounds complex language instructions across diverse scenarios by prioritizing dynamic interactions and spatial reasoning over static appearances. For instance, it accurately identifies actively pushing rugby players and isolates vehicles navigating auxiliary roads. Furthermore, ReaTrack excels at deducing implicit social roles and behavioral semantics. It seamlessly distinguishes yoga students from an instructor, identifies non participants in performances based on attire and positioning, and comprehends complex interactions such as tracking a visually impaired pedestrian alongside a helper. Finally, by analyzing behavioral persistence, such as a subject interacting with a door handle while others depart, the model successfully infers abstract concepts like vehicle ownership. These robust visualizations underscore the framework's remarkable capacity to map implicit reasoning processes into highly precise spatiotemporal trajectories.

\begin{figure}[t]
\setlength{\abovecaptionskip}{0.6mm}
\centering
    \includegraphics[width=1.0\linewidth]{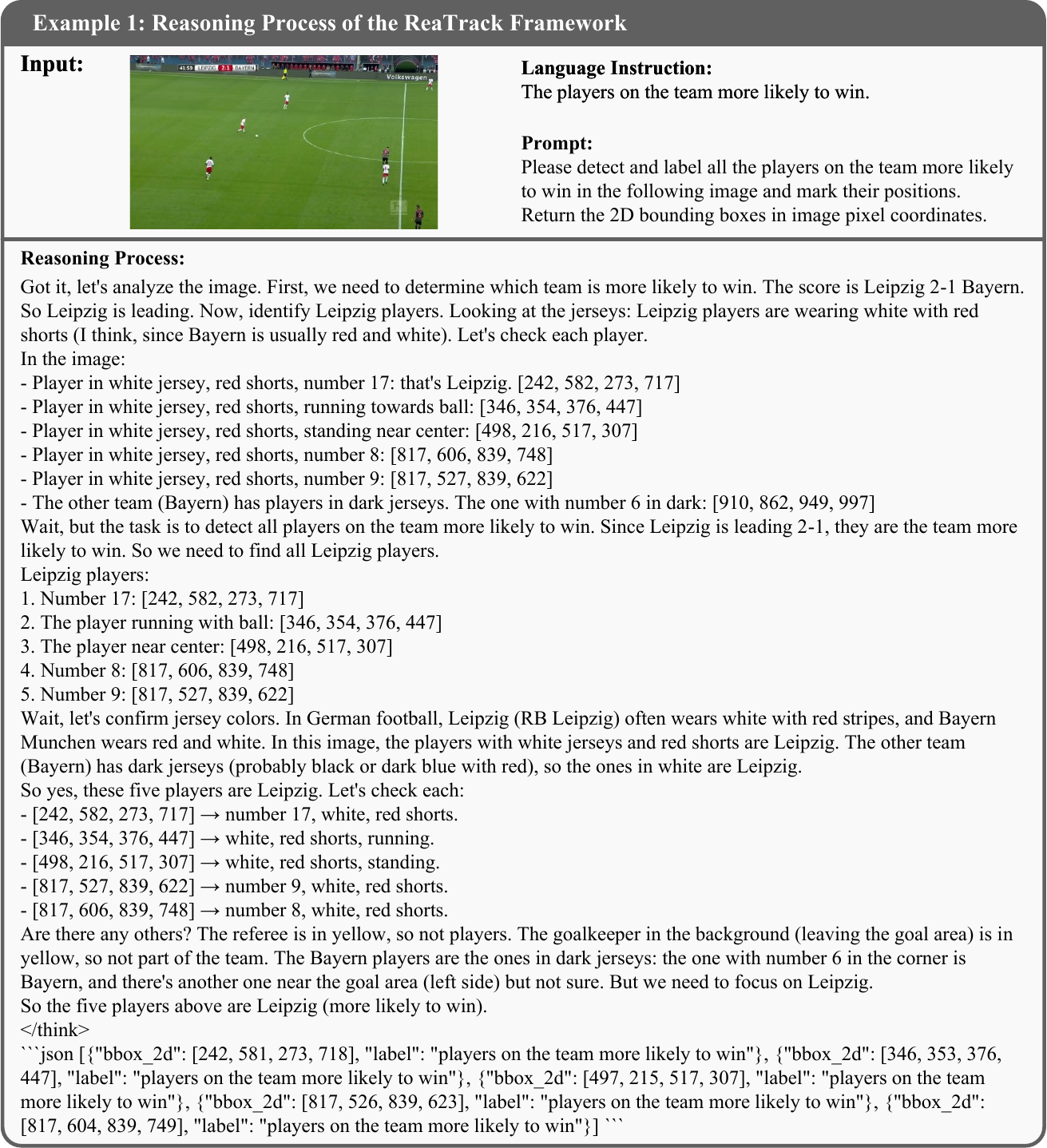}
    \caption{\textbf{Reasoning Process} of the ReaTrack Framework.}
    \label{fig:Appendix Example 1: Reasoning Process.}
\vspace{-2pt}
\end{figure}

\subsection{ReaTrack Reasoning Process Visualization}
\label{subsec:ReaTrack Reasoning Process Visualization}

We provide detailed visualizations of the reasoning process in the ReaTrack framework on the ReaMOT Challenge benchmark to demonstrate how the ReaTrack framework addresses the core requirements of the proposed Reasoning-based Multi-Object Tracking (ReaMOT) task. Unlike traditional tracking tasks that rely on explicit category names, ReaMOT requires models to identify and track targets that satisfy implicit constraints via logical reasoning. As illustrated in Fig.~\ref{fig:Appendix Example 1: Reasoning Process.}, a scenario involving contextual understanding, the model tackles a complex instruction to identify players on the team more likely to win. The task involves implicit constraints that cannot be resolved by visual recognition alone. The framework first analyzes the scoreboard to interpret that Leipzig leads Bayern with a score of two to one and logically deduces that Leipzig is the target team. It then maps this semantic conclusion to visual attributes by recognizing the white jerseys and red shorts of the leading team while successfully excluding non player entities such as the referee and goalkeeper who wear distinct colors.

\begin{figure}[t]
\setlength{\abovecaptionskip}{0.6mm}
\centering
    \includegraphics[width=1.0\linewidth]{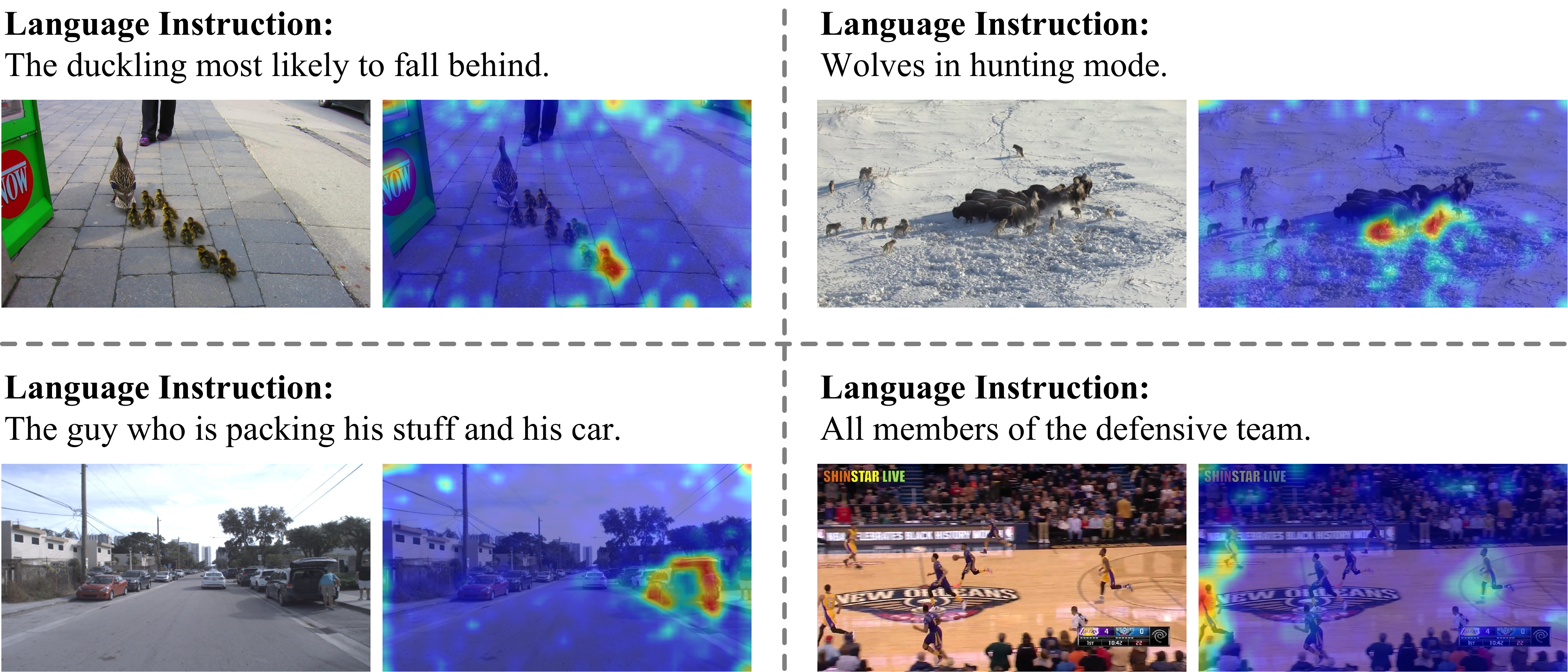} 
    \caption{\textbf{Visualization of attention maps generated by the ReaTrack framework.} The heatmaps explicitly demonstrate how our framework accurately grounds complex, implicit semantic instructions into corresponding visual regions across diverse scenarios. Warmer colors indicate higher attention weights, confirming precise pixel-level semantic alignment.}
    \label{fig:Visualization of attention maps generated by our ReaTrack framework.}
\vspace{-4pt}
\end{figure}

\begin{figure}[t]
\setlength{\abovecaptionskip}{0.6mm}
\centering
    \includegraphics[width=1.0\linewidth]{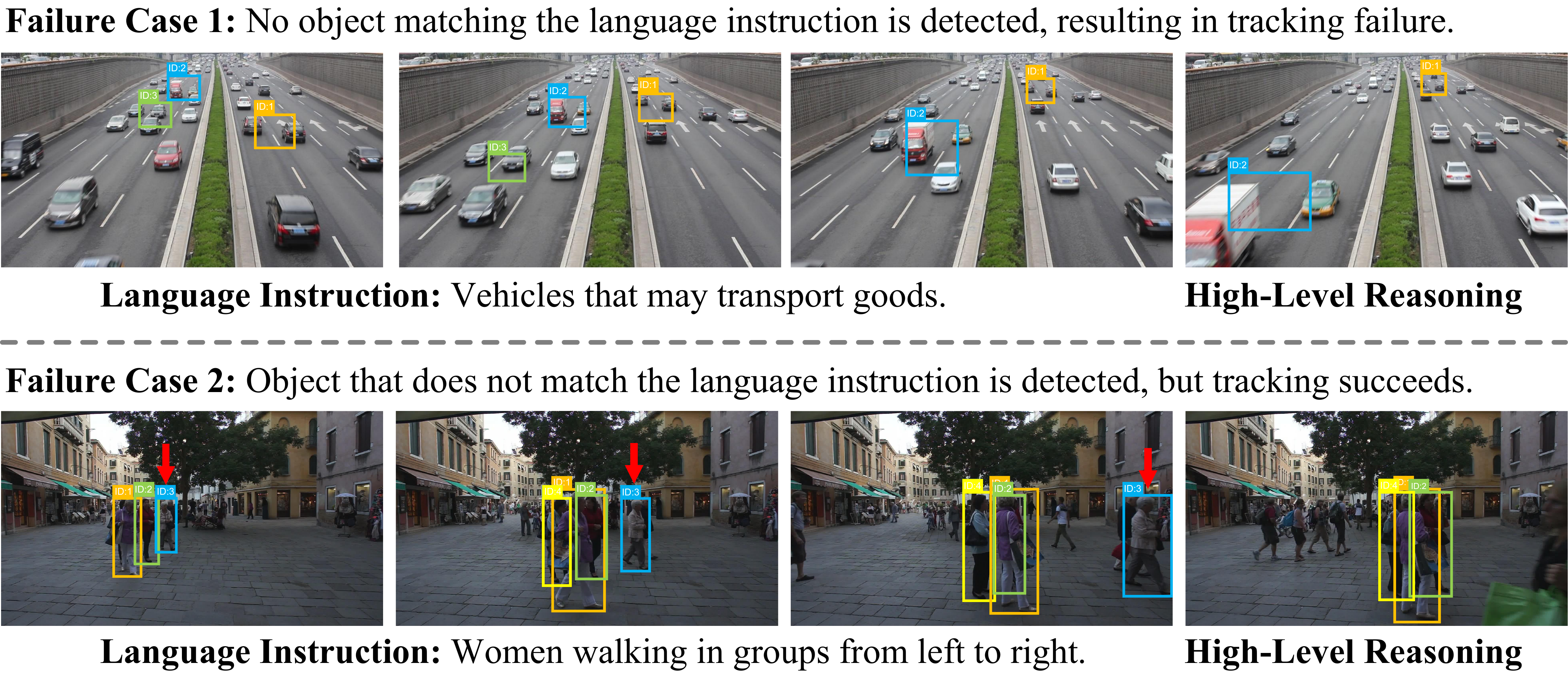} 
    \caption{\textbf{Visualization of representative failure cases.} (1) \textbf{Semantic Omission (False Negative):} The LVLM fails to ground valid targets corresponding to the language instruction, resulting in missed trajectories. (2) \textbf{Semantic Hallucination (False Positive):} Erroneous detection of an irrelevant object that does not satisfy the linguistic constraints, which is subsequently tracked.}
    \label{fig:Failure Results.}
\vspace{-2pt}
\end{figure}

\subsection{Interpretability of Reasoning-Aware Detection  Attention}

To investigate the interpretability of our Reasoning Aware Detection module, Fig.~\ref{fig:Visualization of attention maps generated by our ReaTrack framework.} visualizes spatial attention maps generated by ReaTrack. These visualizations explicitly illustrate the LVLM capacity to ground complex language instructions into specific visual regions. The four presented scenarios highlight entirely different dimensions of cognitive reasoning:

(1) \textbf{Top-Left (Predictive Reasoning):} Instructed to locate the duckling “most likely to fall behind”, the model successfully grounds the target by concentrating its highest attention weights exclusively on the specific individual trailing at the extreme rear of the group. This confirms a strong capacity for spatial sequence analysis and future state prediction.

(2) \textbf{Top Right (Behavioral State Understanding):} Tasked with identifying wolves in “hunting mode”, the attention map precisely activates on predators actively surrounding the prey, correctly ignoring other animals idling in the background. This demonstrates an advanced understanding of dynamic behavioral states over mere static species recognition.

(3) \textbf{Bottom-Left (Human-Object Interaction):} Highlighting a complex interaction, the model concentrates its attention on the person physically engaging with the open trunk. It links the visual sequence of transferring items to the semantic concept of “packing”, establishing a precise spatial ownership.

(4) \textbf{Bottom-Right (Rule-Based Tactical Reasoning):} In a competitive sports scenario requiring the identification of the “defensive team”, the attention heatmap selectively highlights multiple players performing the defensive role. The model successfully deduces this based on their spatial positioning relative to the basket and the opposing team, showcasing robust group-level visual grounding.

Collectively, these heatmaps prove that ReaTrack transcends global context matching, achieving fine grained pixel level semantic alignment to bridge the cognitive gap between implicit linguistic logic and explicit visual evidence.

\begin{figure}[t]
\setlength{\abovecaptionskip}{0.6mm}
\centering
    \includegraphics[width=1.0\linewidth]{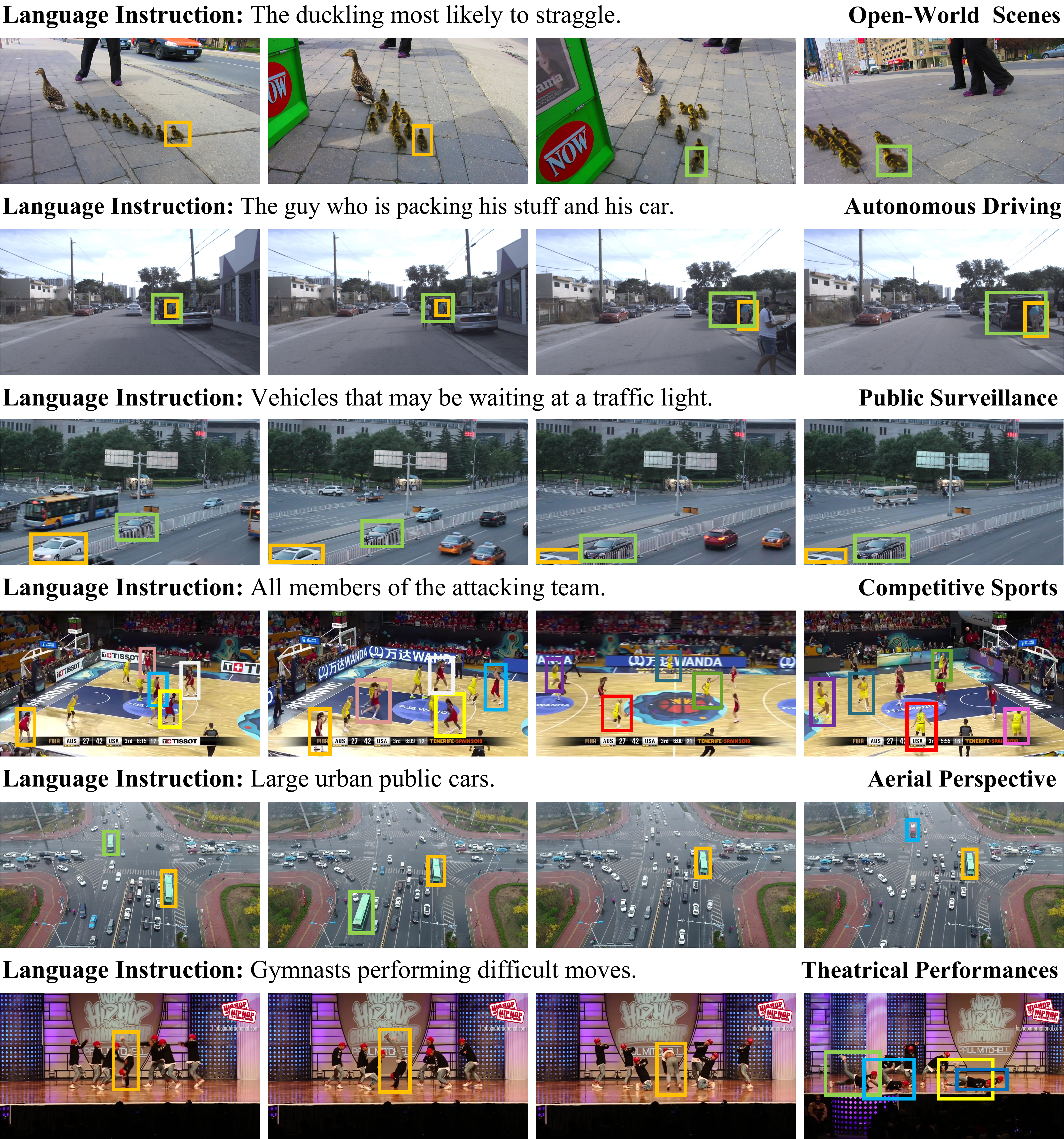}
    \caption{\textbf{Representative examples} of language instructions (High-Level Reasoning) with their ground truth in the ReaMOT Challenge benchmark.}
    \label{fig:ground truth.}
\vspace{-2pt}
\end{figure}

\subsection{Analysis of Failure Cases}
As illustrated in Fig.~\ref{fig:Failure Results.}, our qualitative analysis reveals that tracking errors primarily originate from insufficient semantic reasoning during the per-frame detection phase rather than temporal association inaccuracies. In the first scenario involving semantic omission, the LVLM struggles to map the abstract instruction (“transport goods”) to relevant visual features, resulting in persistent false negatives and completely missed trajectories. In the second scenario, the framework exhibits semantic hallucination. Although the temporal association module maintains highly stable tracking continuity, the underlying detector erroneously grounds an object that explicitly violates the linguistic constraints (such as tracking an individual who is neither in a group nor walking in the specified direction). These representative examples conclusively demonstrate that the current performance bottleneck does not lie in the temporal tracking logic. Instead, it resides in the reasoning engine's capacity to flawlessly translate complex, multi-layered semantic instructions into precise visual bounding boxes.

\subsection{Instructions and Ground Truth in the ReaMOT Challenge}
\label{subsec:Instructions and Ground Truth in the ReaMOT Challenge}

Fig.~\ref{fig:ground truth.} illustrates the language instructions and their corresponding ground truth annotations in the ReaMOT Challenge benchmark. These examples underscore that ReaMOT extends beyond traditional visual recognition, requiring models to bridge the gap between explicit visual cues and implicit semantic meanings. The benchmark encompasses a broad spectrum of diverse reasoning challenges. The cases highlighted here represent a subset of this overarching complexity, illustrating capabilities such as tactical analysis, causal inference, physical interaction understanding, and behavioral prediction. The specific reasoning challenges associated with each showcased example are analyzed below:

\textbf{Reasoning Challenges of Row 1:} This instruction requires predictive behavioral reasoning based on spatial and temporal cues. The model cannot simply detect all ducklings but must evaluate the relative spatial positioning and motion dynamics of the entire group. Identifying the individual “most likely to straggle” demands inferring future states based on the current observation that a specific target is trailing significantly behind the mother and the rest of the brood.

\textbf{Reasoning Challenges of Row 2:} The challenge here involves understanding complex human object vehicle interactions. The model must go beyond basic pedestrian detection to analyze the specific pose and dynamic actions of the person interacting with the open trunk of a vehicle. It requires linking the visual sequence of moving items to the semantic concept of “packing”, while establishing a spatial ownership relationship between the specific person and the target car.

\textbf{Reasoning Challenges of Row 3:} This scenario requires causal inference grounded in common traffic rules. The traffic light itself might not be clearly visible within the camera's field of view. The model must infer the “waiting” state by observing the static nature of a specific group of vehicles halted at a junction line, while perceiving the orthogonal flow of cross traffic, thereby inferring the unseen regulatory constraint causing the target vehicles to stop.

\textbf{Reasoning Challenges of Row 4:} Identifying the “attacking team” demands rule based tactical reasoning. The model must first locate the ball or determine active possession based on player actions. Using prior knowledge of basketball rules to define the attacking role, the model must then generalize this abstract semantic label to all players sharing the same visual attribute (uniform color), regardless of their individual spatial locations or momentary poses on the court.

\textbf{Reasoning Challenges of Row 5:} The challenge lies in mapping abstract functional categories to extreme visual viewpoints. The phrase “Large urban public cars” implicitly refers to buses. The model must rely on prior knowledge linking public transport to specific physical characteristics like elongated rectangular shapes and larger relative sizes, and then recognize these geometric attributes from a top down aerial perspective where fine grained visual details are lost.

\textbf{Reasoning Challenges of Row 6:} This description requires common sense reasoning regarding human biomechanics and physical exertion. Identifying “difficult moves” necessitates distinguishing standard upright postures from complex acrobatic actions. The model must evaluate the pose extremity, balance state, and implied physical effort of each performer to isolate the specific individuals executing flips or demanding freezes from the rest of the stationary group.


\section{Conclusion}
\label{sec:Conclusion}

In this work, we identify the fundamental limitations of existing Referring Multi-Object Tracking (RMOT) paradigms when handling complex reasoning scenarios. To bridge this gap, we propose \textbf{ReaMOT}, a novel task that requires models to identify and track targets based on implicit logical constraints. To advance research in this direction, we establish the \textbf{ReaMOT Challenge}, a comprehensive benchmark featuring hierarchical reasoning levels and diverse scenes, providing a rigorous evaluation platform for the community. Furthermore, we propose \textbf{ReaTrack}, a strong training-free baseline. By decoupling high-level cognitive localization from low-level physical motion, ReaTrack dynamically aligns the semantic detections of a Thinking-variant LVLM with the temporal propagation of SAM2. Extensive experiments demonstrate that our framework achieves \textbf{state-of-the-art} performance, validating the effectiveness of this decoupled design. We hope the ReaMOT task and benchmark will inspire future exploration toward more intelligent video perception systems.

Despite the promising results, several challenges remain. First, ReaTrack's reliance on large vision-language models introduces computational overhead and potential hallucination risks. Second, decoding highly abstract semantic cues, such as causal intent and complex social roles, remains formidable for current architectures. Future work will focus on exploring end-to-end trainable models that jointly optimize reasoning and tracking. Additionally, developing lightweight distillation techniques for real-time deployment, alongside finer-grained metrics to evaluate the correctness of intermediate reasoning chains rather than trajectory accuracy alone, represent critical directions for advancing this field.


{\small
\bibliographystyle{IEEEtran}
\bibliography{reference}
}

\end{document}